\documentclass{article}

\PassOptionsToPackage{numbers, compress}{natbib}

\usepackage[preprint]{neurips_2023_ml4ad}

\usepackage[utf8]{inputenc} 
\usepackage[T1]{fontenc}    
\usepackage{hyperref}       
\usepackage{url}            
\usepackage{booktabs}       
\usepackage{amsfonts}       
\usepackage{nicefrac}       
\usepackage{microtype}      
\usepackage{xcolor}         

\usepackage[linesnumbered,ruled]{algorithm2e}
\usepackage{subcaption}
\usepackage{pgfplots}
\usepackage{tikz}
\usepackage{rotating}

\usepackage[acronym]{glossaries}
\makeglossaries
\newacronym{obb}{OBB}{Oriented Bounding Box}
\newacronym{vits}{ViTs}{Vision Transformers}

\title{Explainable Multi-Camera 3D Object Detection with Transformer-Based Saliency Maps}

\author{%
    Till~Beemelmanns, Wassim~Zahr, Lutz~Eckstein\\
    RWTH Aachen University\\
	\texttt{\{till.beemelmanns, lutz.eckstein\}@ika.rwth-aachen.de}\\
    \texttt{wassim.zahr@rwth-aachen.de}
}

\begin{document}

\maketitle

\begin{abstract}
Vision Transformers (ViTs) have achieved state-of-the-art results on various computer vision tasks, including 3D object detection. However, their end-to-end implementation also makes ViTs less explainable, which can be a challenge for deploying them in safety-critical applications, such as autonomous driving, where it is important for authorities, developers, and users to understand the model's reasoning behind its predictions. In this paper, we propose a novel method for generating saliency maps for a DetR-like ViT with multiple camera inputs used for 3D object detection. Our method is based on the raw attention and is more efficient than gradient-based methods. We evaluate the proposed method on the nuScenes dataset using extensive perturbation tests and show that it outperforms other explainability methods in terms of visual quality and quantitative metrics. We also demonstrate the importance of aggregating attention across different layers of the transformer. Our work contributes to the development of explainable AI for ViTs, which can help increase trust in AI applications by establishing more transparency regarding the inner workings of AI models.
\end{abstract}

\section{Introduction}
\label{sec:introduction}
Vision Transformers (ViTs) have made significant strides in the field of computer vision, achieving state-of-the-art results on a wide range of scene understanding tasks for automated driving. Transformer-based 3D object detection was successfully applied for LiDAR \cite{zhou2022centerformer}, multi-view cameras \cite{Wang2022, liu2022petr, li2022bevformer, Doll2022} or for multi-modal input \cite{yan2023cross}. The idea is that the transformers' attention mechanism is able to capture a global understanding of the scene and therefore generates more accurate detections and also eliminates the need for handcrafted postprocessing or object fusion steps. This end-to-end approach offers many advantages, but it also presents a challenge in terms of explainability \cite{9565103}. This challenge becomes particularly crucial when considering the deployment of ViTs in safety-critical applications like autonomous driving. In such scenarios, it is essential for authorities, developers, and users to have a clear understanding of the model's reasoning behind its predictions.

The generation of saliency maps is a commonly employed method for enhancing the explainability of DNNs. Saliency maps reveal the most critical areas that are relevant for the model's output. For CNN networks, many approaches exists that either use expensive perturbation-based methods \cite{Ribeiro2016, Fong2017, Fong2019, Petsiuk2018}, use the networks' gradients \cite{Simonyan2013, Selvaraju2017} or apply manual propagation rules \cite{Bach2015} to derive the relevancy maps. Transformer-based approaches use attention maps as a source of explainability. This method has been applied to NLP tasks \cite{Abnar2020}, image classification \cite{Chefer2021} and object detection \cite{Chefer2021_mm, Abeloos2022}. However, there remains a gap in the literature when it comes to exploring an approach for 3D object detection in the context of AD with multi-sensory input.

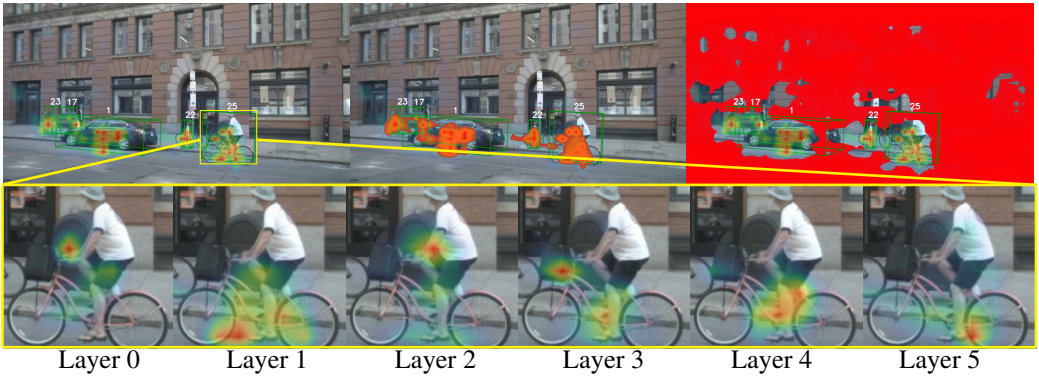
\begin{figure}[h]
    \centering
    
    \begin{tikzpicture}
    
    \node[anchor=south west] at (0, 0) {\includegraphics[width=0.33\textwidth]{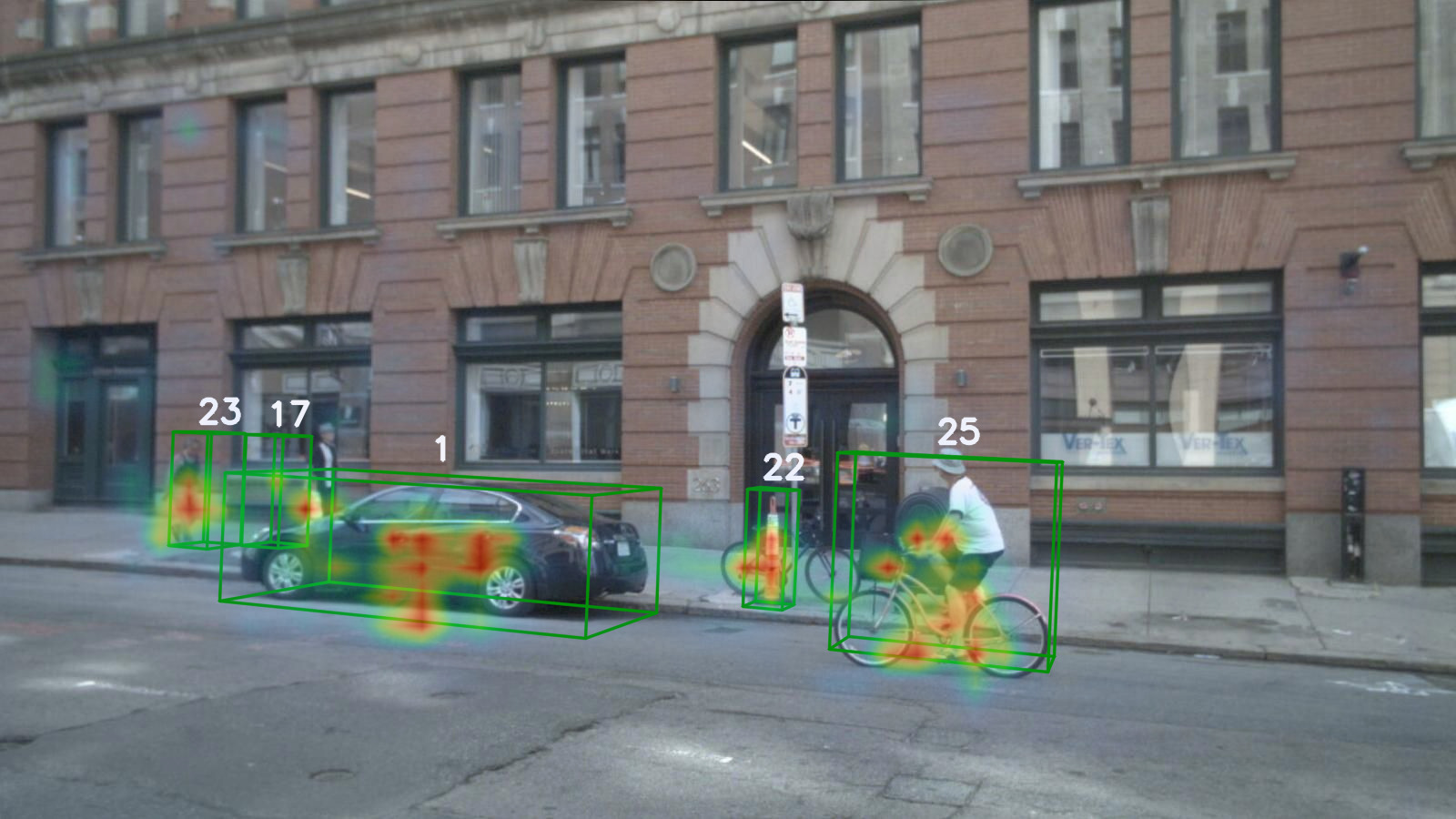}};
    
    \node[anchor=south west] at (0.33\textwidth, 0) {\includegraphics[width=0.33\textwidth]{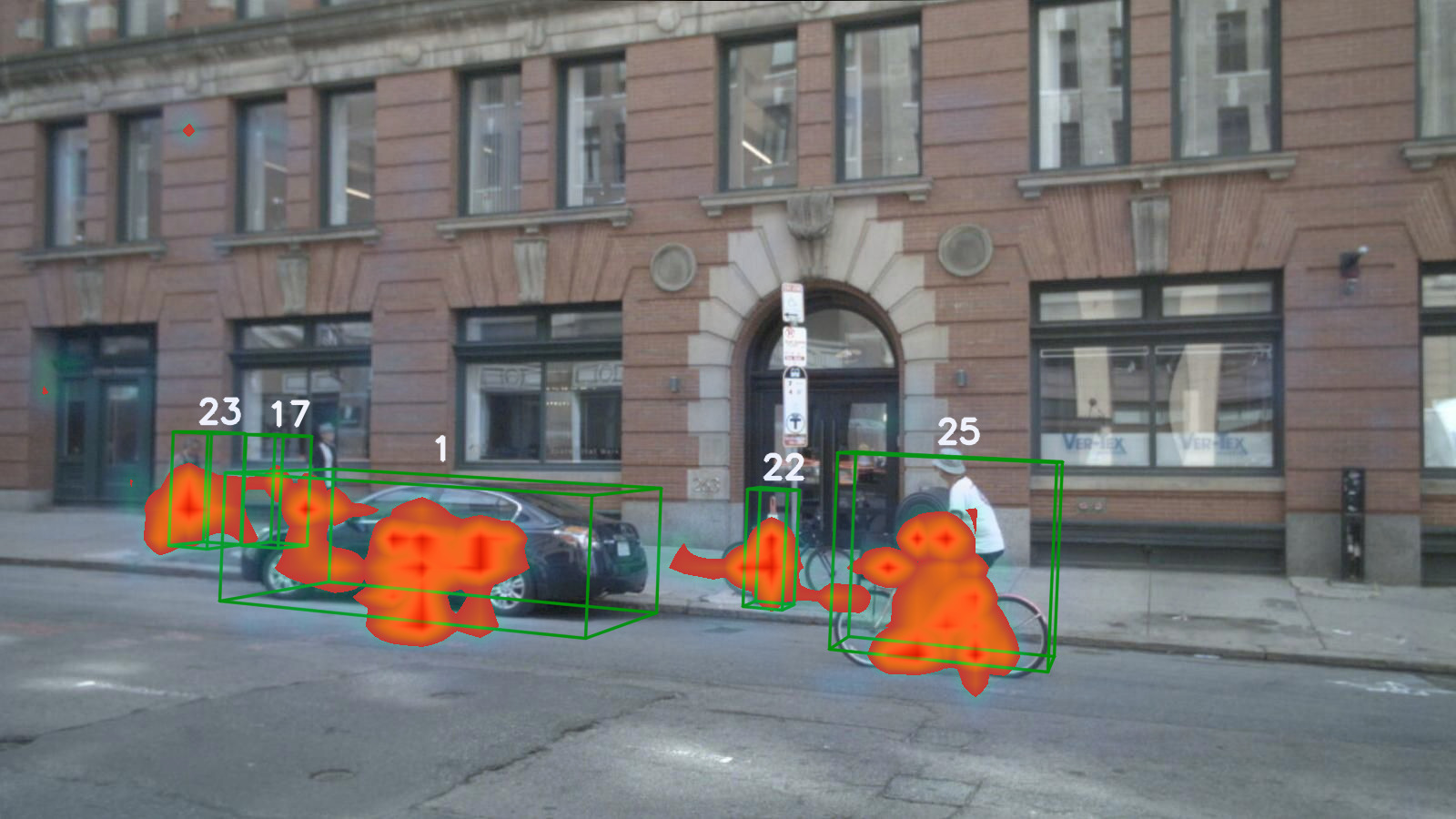}};
    
    \node[anchor=south west] at (0.65\textwidth, 0) {\includegraphics[width=0.33\textwidth]{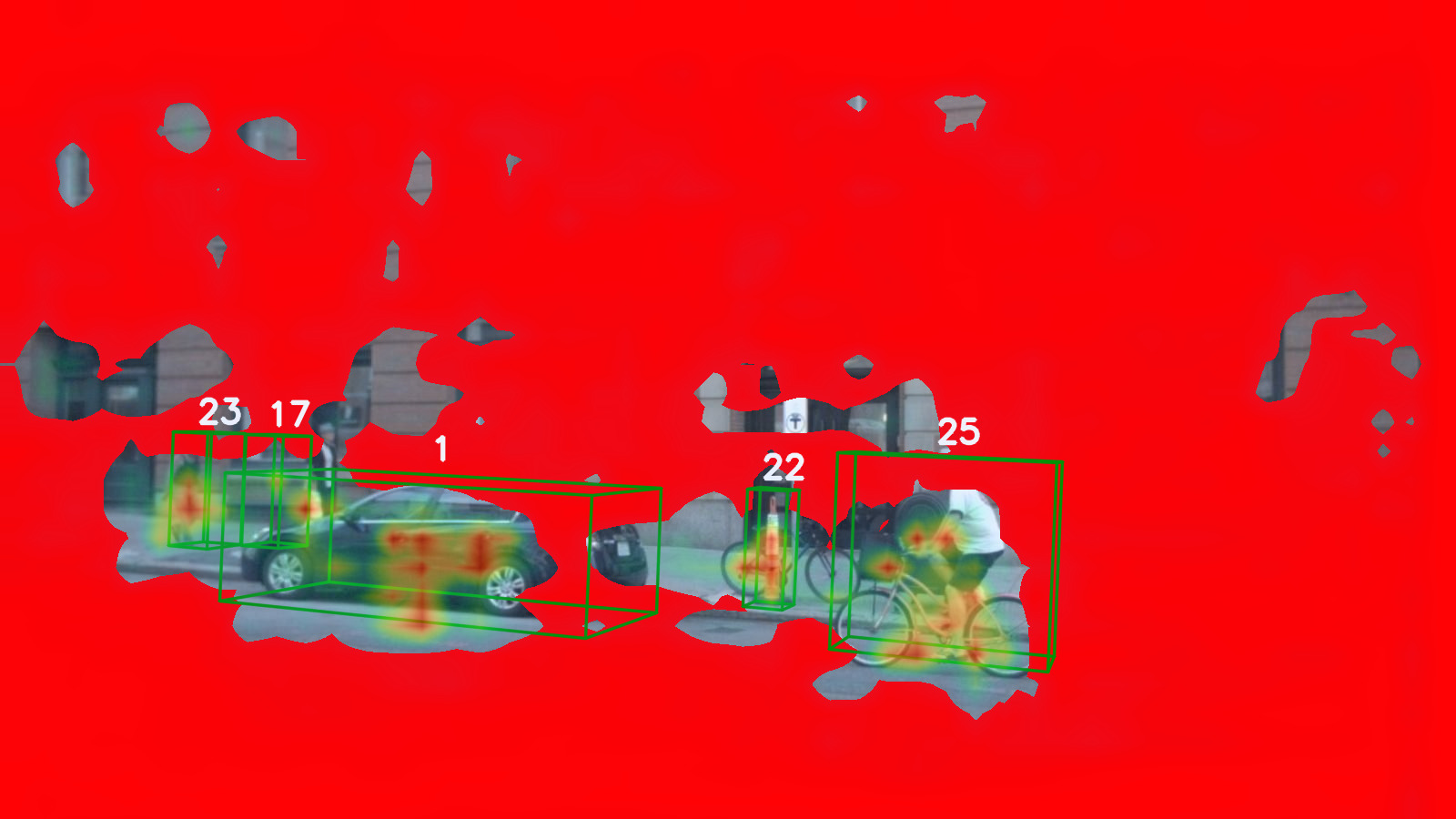}};
    
    \node[anchor=south west] at (0, -2) {\includegraphics[width=0.165\textwidth]{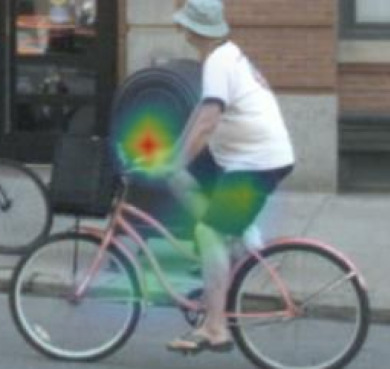}};
    \draw (0.10\textwidth, -2.05) node {Layer 0};

    \node[anchor=south west] at (0.162\textwidth, -2) {\includegraphics[width=0.165\textwidth]{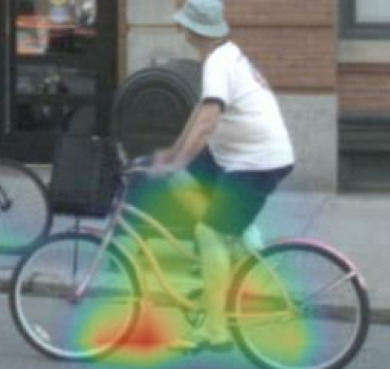}};
    \draw (0.26\textwidth, -2.05) node {Layer 1};
    
    \node[anchor=south west] at (0.326\textwidth, -2) {\includegraphics[width=0.165\textwidth]{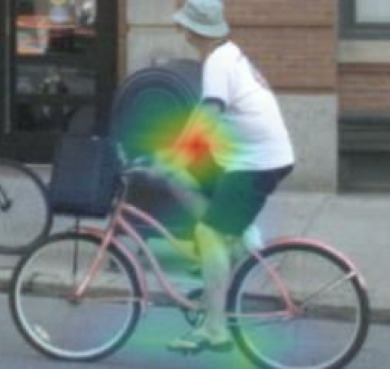}};
    \draw (0.42\textwidth, -2.05) node {Layer 2};
    
    \node[anchor=south west] at (0.490\textwidth, -2) {\includegraphics[width=0.165\textwidth]{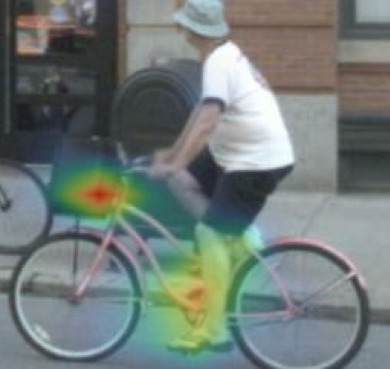}};
    \draw (0.58\textwidth, -2.05) node {Layer 3};

    \node[anchor=south west] at (0.654\textwidth, -2) {\includegraphics[width=0.165\textwidth]{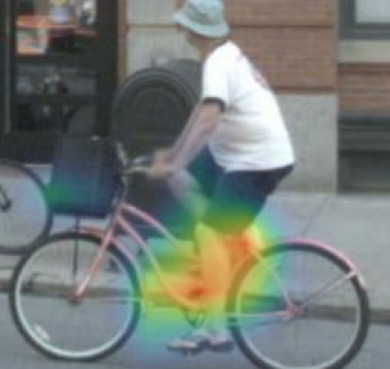}};
    \draw (0.74\textwidth, -2.05) node {Layer 4};
    
    \node[anchor=south west] at (0.818\textwidth, -2) {\includegraphics[width=0.165\textwidth]{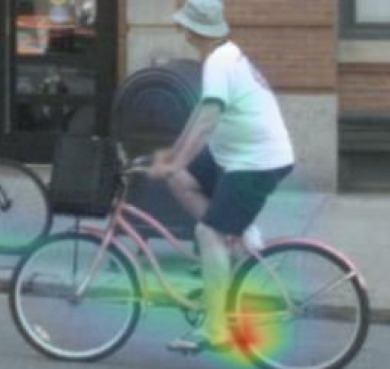}};
    \draw (0.90\textwidth, -2.05) node {Layer 5};

    \draw[draw=yellow] (2.74,0.58) rectangle ++(0.75,0.71);
    \draw[yellow,line width=1pt,-] (2.74, 0.9) -- (0.1, 0.30);
    \draw[yellow,line width=1pt,-] (3.49, 0.9) -- (13.85, 0.30);
    \draw[yellow,line width=1pt,-] (0.11, 0.30) -- (13.85, 0.30);
    \draw[yellow,line width=1pt,-] (0.11, 0.30) -- (0.11, -1.86);
    \draw[yellow,line width=1pt,-] (0.11, -1.86) -- (13.85, -1.86);
    \draw[yellow,line width=1pt,-] (13.85, -1.86) -- (13.85, 0.30);

    \end{tikzpicture}
    
    \caption{This work aims at generating saliency maps as visual explanations for a multi-camera transformer-based 3D object detector. We perform extensive perturbation tests to evaluate raw attention and gradient-based methods. We conduct positive (top-middle) and negative (top-right) perturbation tests, where certain parts of the six input camera images are masked (red) to measure the effectiveness of each approach. We found that it is necessary to aggregate attention across all transformer layers (bottom) to produce saliency maps that are comprehensive.}
    \label{fig:overview}
\end{figure}

In this work, we propose the first saliency map generation approach for a multi-camera transformer model, aiming to enhance our understanding of the model's behavior and provide valuable insights into which regions of input images are most influential in determining object detection results. We propose a simple and effective method to generate saliency maps which outperforms gradient-based methods. We validate the effectiveness of our approach through comprehensive perturbation tests. See Figure \ref{fig:overview} for an overview. Our main contributions are the following:
\begin{itemize}
    \item We propose a novel method for generating saliency maps for a transformer-based multi-camera 3D object detector. This method is based on the raw cross-attention and is more efficient than gradient-based methods.
    \item We demonstrate the importance of aggregating attention across different layers of the transformer by using extensive perturbation test. This aggregation helps to produce comprehensive and informative saliency maps.
\end{itemize}

\section{Related Work}
\label{sec:related_work}
\subsection{Explainable AI}

\emph{Transparent models} or white-box models can be easily interpreted by humans straight from their structure \cite{Arrieta2020}. For example, in linear regression models, the weight assigned to each feature directly indicates its importance. Similarly, decision trees provide a clear path from the root node to the leaf nodes. However, applying those models to high-resolution image data is usually not efficient.


\emph{Post-hoc} explanation methods are used to understand how more complex models, such as DNNs, SVM, and CNNs, make their predictions \cite{Arrieta2020}. For example, Grad-CAM \cite{Selvaraju2017} is a post-hoc explanation method for CNNs, which uses gradients of the target class in the final convolutional layer to create a heatmap highlighting important regions in the input image that contributed to the prediction.

\emph{Model-specific} explainability techniques are specific to certain models, using their unique structures for explanation purposes. For example, a CNN can be interpreted using Layer-wise Relevance Propagation (LRP). This method pushes the output decision backward through the layers to the input, giving a relevance score to each input feature \cite{Bach2015}. Utilizing LRP can be challenging because it requires a specific implementation for each architecture.

\emph{Model-agnostic} techniques do not depend on detailed knowledge of the model's structure. Methods like Local Interpretable Model-Agnostic Explanations (LIME) \cite{Ribeiro2016} and SHapley Additive exPlanations (SHAP) fall under this group \cite{Lundberg2017}. They approximate the local decision boundary of any model and can assign scores to the importance of input features, but they are also computationally inefficient.

\subsection{Saliency Maps}
Most explainability methods in Computer Vision generate \emph{saliency maps}, revealing the most critical or salient areas for image comprehension \cite{Simonyan2013, Selvaraju2017, Chefer2021, Chefer2021_mm, Ghiasi2022}.

\emph{Perturbation-based} methods pinpoint key features or regions in input data by systematically altering them and monitoring the effect on the prediction outcome by the model. LIME \cite{Ribeiro2016} is a local approximation to the decision boundary of complex classification models. Fong et al. \cite{Fong2017, Fong2019} proposed a framework to unveil explanations by identifying the specific image fragment that affects a classifier's decision. Randomized Input Sampling for Explanations (RISE) \cite{Petsiuk2018} generates  saliency maps by randomly masking portions of an image and aggregating the resulting outputs. However, these methods entail expensive data perturbation and multiple model forward passes, hindering their efficient use for an application in real-time scenarios with a multi-camera setup.


\emph{Gradient-based} methods determine the output category's gradient concerning the input image. This gradient can be visualized and interpreted as a saliency map. Grad-CAM \cite{Selvaraju2017} distinguishes between classes based on these visual explanations, which is particularly important for object detection. For ViTs, Grad-CAM generates class-specific visualizations by multiplying the attention maps with the corresponding gradients \cite{Chefer2021, Chefer2021_mm}.

\emph{LRP} \cite{Bach2015} distributes the model's prediction back to its input features by traversing the network backwardly. Each neuron's relevance in a layer is calculated based on its contribution to the neurons in the subsequent layer. This relevance propagation continues layer by layer until it reaches the input layer. However, applying LRP to transformers has proven challenging due to the complex interaction patterns within the attention mechanism, and its model-specific approach \cite{Abnar2020}.

\emph{Attention} mechanism in transformer models assigns scores that reflect the significance of various segments within a sequence of input tokens, allowing the model to focus on the most relevant input. \emph{Attention Rollout} interprets these attention scores across layers to give a more comprehensive understanding of a transformer's decision-making process \cite{Abnar2020, Chefer2021}. However, this method is computational inefficient for large-scale evaluations and it is unable to distinguish between positive and negative contributions to a decision. Furthermore, it's application is limited to the self-attention mechanism, whereas modern 3D object detectors usually integrate cross-attention as well.

\emph{Gradient Rollout}, a method proposed by Chefer et al. \cite{Chefer2021_mm}, considers attention mechanisms across all layers to generate a \emph{relevancy map}, and each layer contributes to this map following a specific set of rules. While its effectiveness has been evaluated for an encoder-decoder transformer for 2D object detection, its applicability to decoder-only architectures mainly used in 3D object detection remains unexplored. Additionally, it requires gradient computation which introduces a computational overhead.  

\subsection{2D Object Detection}
Early object detection relied on two-step approaches, where \emph{region proposals} were first generated and then classified with SVM-based classifiers \cite{Dalal2005}. RCNNs (Region-based CNNs) made use of CNNs to extract features for region proposal and classification \cite{Girshick2014}. Fast RCNN \cite{Girshick2015} and Faster RCNN \cite{Ren2015}, improved inference speed and accuracy of these region-based methods.

\emph{YOLO} went on to redefine object detection as a regression problem to predict spatially distinct bounding boxes and class probabilities in a single pass \cite{Redmon2016}. It divides the input image into a grid structure where each cell predicts multiple bounding boxes and class probabilities. 
%

\emph{DetR} introduced a different approach to object detection by utilizing the transformer architecture \cite{Carion2020}. DetR views object detection as a \emph{set prediction problem} and it learns to directly predict a fixed-size set of bounding boxes and object classes for the entire image in one go, eliminating the need for region proposal, handcrafted anchor boxes and Non-Maximum Suppression (NMS). 

\subsection{Camera-based 3D Object Detection}
\emph{Monocular} image-based approaches to 3D object detection rely on a single camera as the sensory input. Wang et al. proposed FCOS3D \cite{Wang2021}, which is based on FCOS \cite{Tian2019}, a method to transform 3D targets with a center-based paradigm, avoiding any necessary 2D detection or 2D-3D correspondence priors. In FCOS3D, the 3D targets are projected onto the 2D image plane, resulting in a projected center point.

\emph{Pseudo-LiDAR} approaches aim to bridge the gap between 3D object detection using LiDAR and monocular or stereo images. Wang et al. proposed a method to transform a camera depth estimation into a point cloud format, called the pseudo-LiDAR \cite{Wang2019}. Then existing LiDAR-based 3D object detection pipelines where applied on the 3D pseudo representation. Many other approaches improved this idea and achieved impressive results \cite{DBLP:journals/corr/abs-1903-09847, DBLP:journals/corr/abs-2004-03080, DBLP:journals/corr/abs-1903-11444}.

Similar to 2D object detection, \emph{transformer-based} architectures have been adopted to 3D object detection, leveraging the ability of transformers to capture long-range dependencies from multiple-camera streams. DETR3D \cite{Wang2022} extracts 2D features from multiple camera images and uses 3D object queries to index these features. This is achieved by linking 3D positions to multi-view images using camera transformation matrices. Similar approaches introduce improved query, key and value design. For example, PETR \cite{liu2022petr} uses a special Position Embedding TRansformation (PETR) for the same multi-camera setup to encode the position information of 3D coordinates into image features, producing the 3D position-aware features. SpatialDETR \cite{Doll2022} introduced a geometric positional encoding which incorporated view geometry to explicitly consider the extrinsic and intrinsic camera setup. BEVFormer \cite{li2022bevformer} creates a discretised 3D world in the bird's eye view perspective and considers each grid as a query location. 



\subsection{Transformer}


The decoder-only transformer model is composed of a stack of layers. Each layer consists of two modules: a \emph{multi-head attention} mechanism and a position-wise \emph{FFN}. The attention mechanism is deployed usually two times: for the \emph{self-attention} and for the \emph{cross-attention} mechanisms. These mechanisms allow the model to assess the relevance of different input tokens while generating the output. The input of the attention mechanism consists of three learned vectors: \emph{query} ($Q$), \emph{key} ($K$), and \emph{value} ($V$). 

The transformations for each head $i=1,2,\ldots,n_h$ of the \emph{multi-head attention} are given by
\begin{equation}
Q_i = Q \cdot W_{Qi},\;\;\;\; K_i = K \cdot W_{Ki},\;\;\;\; V_i = V \cdot W_{Vi}
\end{equation}
where $W_{Qi}$, $W_{Ki}$, and $W_{Vi}$ are the learned linear transformations for the head. These transformations project the input vectors into the transformer embedded space, allowing each head to focus on a different aspect of the input data. 
Suppose the input query vector $Q \in \mathbb{R}^{n_q \times m}$, and the transformer model has embedded dimension $d$. Every attention computation is executed within a query size $s$, which is given by the formula $s=d/n_h$, where $n_h$ represents the number of attention heads. For each head, the learned linear transformation is denoted as $Q_{i} \in {R}^{m \times s}$. Consequently, the query vector $Q_i \in {R}^{n_q \times s}$, i.e. is projected into the space with dimension $s$.

Then, the scaled dot-product attention \cite{Vaswani2017} for each head $i$ is given by
\begin{equation}\label{eq:attention}
A(Q_i, K_i, V_i) = \mbox{Softmax} \left( \frac{Q_i K_i^T}{\sqrt{d}} \right) V_i,
\end{equation}
which allows the model to assign different levels of importance to different parts of the input sequence. 

The final step in the multi-head attention mechanism involves concatenating the outputs of the $n_h$ attention heads and linearly transforming the result to produce the final output, which will lead to the vectors being projected back into the original embedding space of the transformer
\begin{equation}
\mbox{MultiHead}(Q, K, V) = \mbox{Concat}(\mbox{head}_1, \ldots, \mbox{head}_{n_h})W_O,
\end{equation}
where $\mbox{head}_i = A(Q_i, K_i, V_i)$ and $W_O$ is a learned linear transformation which combines the outputs of all the heads, allowing the model to capture a wide range of information from the input data.

\subsection{SpatialDETR}
\begin{figure}[!htbp]
  \centering
  \includegraphics[width=0.80\textwidth]{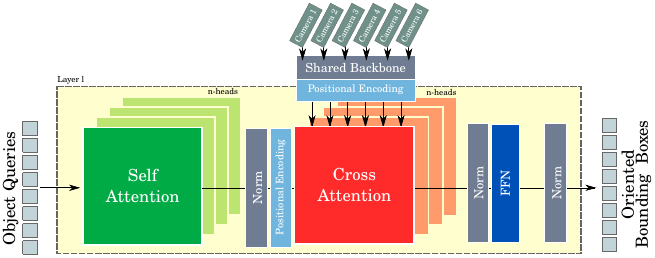}
  \caption{SpatialDETR is a decoder-only transformer architecture. It consists of a query-based self-attention and a cross attention between queries and image features from six cameras \cite{Doll2022}.}
  \label{fig:spatialdetr}
\end{figure}

In this study we examine SpatialDETR \cite{Doll2022}, a transformer-based architecture that takes six multi-view camera images as its input and predicts 3D Oriented Bounding Boxes (OBBs). The images are passed through a shared backbone network, and the resulting representations are then fed into the transformer decoder, as shown in Figure \ref{fig:spatialdetr}. Like other object detectors \cite{liu2022petr, Doll2022, Wang2022, li2022bevformer}, SpatialDETR does not employ an \emph{encoder} for the input to prevent the costly self-attention between the image representation. SpatialDETR follows the design of DETR3D, but additionally introduces a \emph{geometric} positional encoding and a \emph{spatially-aware} attention mechanism, which accounts for the extrinsic spatial information of the multi-camera setup and supports a \emph{global attention} across all cameras, which makes this architecture a particularly good choice to generate global explanations w.r.t. the model's detections.

The detections are generated using $n_q$ object queries which are iteratively refined in each layer of the decoder. Within the decoder layer, \emph{self-attention} is applied across the queries and \emph{cross-attention} between object queries and image features. The final bounding boxes are predicted using a FFN, which transforms each latent object query into an OBB. 

\section{Method}
We use the model's attention layers to produce saliency maps for the interactions in the cross and self-attention. In the following, we discuss different gradient-free and gradient-based computation and propagation rules. The goal of each method is to compute saliency map for each of the input camera images $S \in \mathbb{R}^{n_{c} \times H \times W}$ in the original camera image dimensions $H \times W$.

\subsection{Saliency Map Generation}
Let $n_q$, $n_t$ be number of queries and the number of image tokens per camera respectively. The decoder-only architecture consists of only two types of interactions between the input tokens. The self-attention interaction between the queries $A_{SF}$ and the multi-modal cross-attention between image tokens and queries $A_{CR}$. In the cross-attention module, the queries are interacting with all tokens of all $n_c$ camera images simultaneously. The interactions happen in each of the $n_h$ heads of the multi-head attention for every layer of the $n_l$ transformer decoder layers. During the inference of the model, we aggregate the attention which results in a multi-dimensional cross-attention $A_{CR} \in \mathbb{R}^{n_{c} \times n_{l} \times n_{h} \times n_{q} \times n_{t}}$ and self-attention $A_{SF} \in \mathbb{R}^{n_{l} \times n_{h} \times n_{q} \times n_{q}}$ tensor. As the model is conditioned in a set-to-set fashion, each query $q$ will result in a classification vector $y_q$. Each element of this vector represents the sigmoidal class probability for all considered classes. If none of the class probabilities exceeds a certain threshold, the query will be regarded as background and won't be considered in the computation of the saliency maps.

\subsection{Raw-Attention Methods}
For the first methods, we will consider only the \emph{raw attention} of the cross-attention. We follow \cite{Chefer2021_mm} and define a baseline which takes the cross-attention from the last decoder layer
\begin{equation}
    S = \mathbb{E}_h ( A_{CR}^{n_l} )^+
\end{equation}
where $A_{CR}^{n_l}$ is the last cross-attention map and $\mathbb{E}_h$ denotes an average across heads. We follow \cite{Chefer2021_mm} and remove negative contributions before layer averaging. Different fusion approaches across \emph{layers} of the transformer have not been explored by relevant literature \cite{Chefer2021, Chefer2021_mm}. Hence, we additionally propose a \emph{mean-layer} and a \emph{max-layer} aggregation strategy. The former one computes the mean across all layers and across all heads
\begin{equation}\label{eq:mean}
    S = \mathbb{E}_l ( \mathbb{E}_h( A_{CR} )^+ )
\end{equation}
whereas the latter one filters the maximum values across all layers and heads
\begin{equation}\label{eq:max}
    S = \mbox{max}_l ( \mbox{max}_h( A_{CR} ) ).
\end{equation}

\subsection{Gradient-Based Methods}
For the gradient-based methods we use the \emph{Grad-CAM} \cite{Selvaraju2017} adaptation described in \cite{Chefer2021}. We further adapt the method to the SpatialDETR architecture and compute
\begin{equation}
    S = \mathbb{E}_l ( \mathbb{E}_h ( \nabla A_{CR} \odot A_{CR} )^+ )
\end{equation}
for each layer and for each camera, where $\nabla A_{CR}:= \frac{\partial y_q}{\partial A_{CR}}$ is the cross-attention gradient w.r.t. model output $y_t$. In contrast to \cite{Chefer2021, Chefer2021_mm}, we not only consider the decoder's last layer, but instead also aggregate across all layers. We found that this approach captures more information than only considering the last layer.

We adapted \emph{Gradient Rollout} \cite{Chefer2021_mm}, originally developed for encoder-decoder transformers, to our decoder-only detector. We initialize relevancy maps for the self-attention and cross-attention with 
\begin{equation}
    R_{qq} = \mathbb{I}, \;\; R_{qi} = \emptyset
\end{equation}
and then perform a layer-wise iterative update for the self-attention with
\begin{equation}
    \bar{A}_{SF} = \mathbb{E}_h ( \nabla A_{SF} \odot A_{SF} )^+
\end{equation}
where $ A_{SF} \in \mathbb{R}^{n_q \times n_q}$ and $\mathbb{E}_h$ computes the mean across all heads. Then, we update the the relevancy maps for the self-attention 
\begin{equation}
    R_{qq} =  R_{qq} + \bar{A}_{SF} \cdot R_{qq}
\end{equation}
\begin{equation}
    R_{qi} =  R_{qi} + \bar{A}_{SF} \cdot R_{qi} .
\end{equation}
Next, we aggregate the cross-attentions and update $R_{qi}$
\begin{equation}
    \bar{A}_{CR} = \mathbb{E}_h ( \nabla A_{CR} \odot A_{CR} )^+
\end{equation}
\begin{equation}
    R_{qi} =  R_{qi} + (\bar{R}_{qq})^T \cdot \bar{A}_{CR}
\end{equation}
where $\bar{R}_{qq}$ is the row-wise normalized matrix of ${R}_{qq}$, as described in \cite{Chefer2021_mm}. After passing the last layer of the decoder, the saliency map is obtained by the aggregated cross-attention relevancy map, i.e. $S = {R}_{qi}$. More details on the implementation of the algorithms can be found in Appendix \ref{appendix:algorithms}. Note that all methods require only a few simple hooks in the attention modules. The gradient-based methods need substantially more memory and runtime compared to the raw-attention based methods.

\section{Experiments}
We perform extensive experiments on the nuScenes dataset \cite{Caesar2020}. First, we perform a qualitative visual exploration of the attentions maps to better understand the mechanism of the architecture. Second, we quantitatively evaluate the quality of the saliency maps produced by each method using \emph{positive} and \emph{negative perturbation} tests. Third, we follow \cite{Adebayo2018} and perform a simple sanity check for saliency maps.

\subsection{Visual Exploration} \label{sec:visual_exploration}
\begin{figure}[h]
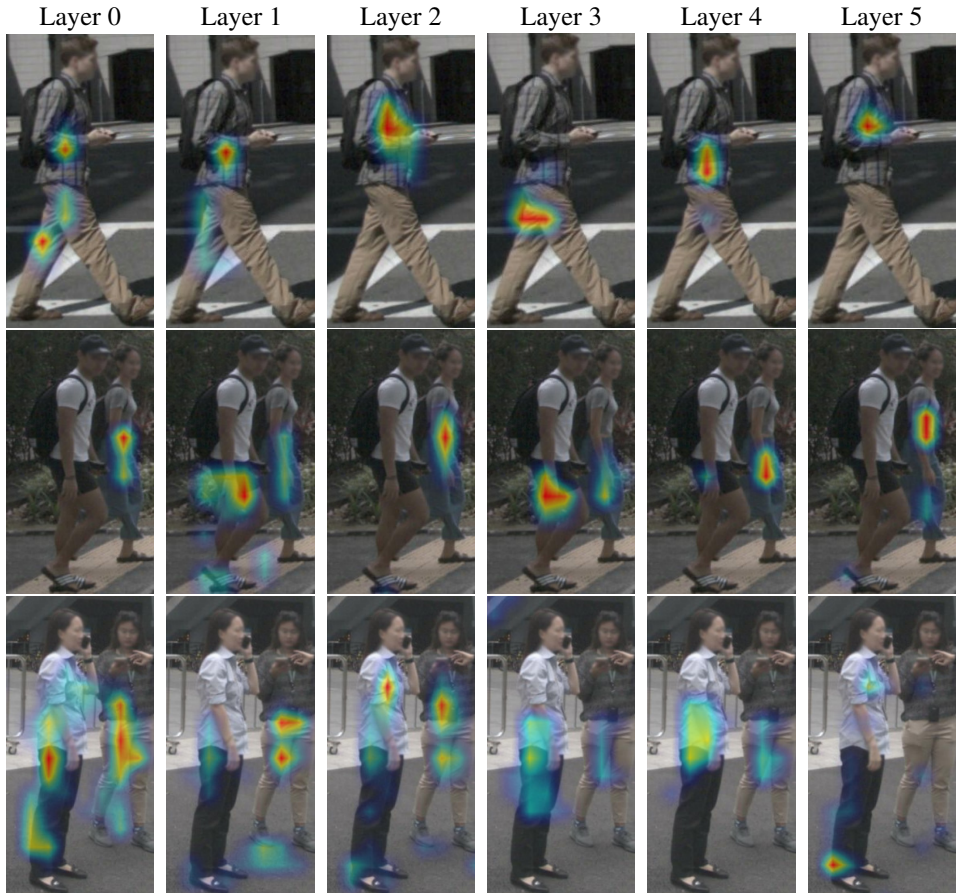

    \centering
    \foreach \i in {0,...,5} {  
        \begin{minipage}{0.14\textwidth}
            \centering
            {Layer \i}  
        \end{minipage}
    }
    \foreach \i in {0,...,5} {  
        \begin{subfigure}[b]{0.14\textwidth}
            \includegraphics[width=1\textwidth]{2873_side/layer_\i}
        \end{subfigure}
    }
    \foreach \i in {0,...,5} {  
        \begin{subfigure}[b]{0.14\textwidth}
            \includegraphics[width=1\textwidth]{5151_side/layer_\i}
        \end{subfigure}
    }
    \vspace{5mm}
    \foreach \i in {0,...,5} {  
        \begin{subfigure}[b]{0.14\textwidth}
            \includegraphics[width=1\textwidth]{271_double/layer_\i}
        \end{subfigure}
    }
\caption{Raw cross-attention $\mathbb{E}_h( A_{CR} )^+$ for a single query shown for all six decoder layers for different pedestrian. Each layer shifts it's focus on a different conceptual part of the pedestrian.}
\label{fig:peds_side}
\end{figure}

\begin{figure}[h]
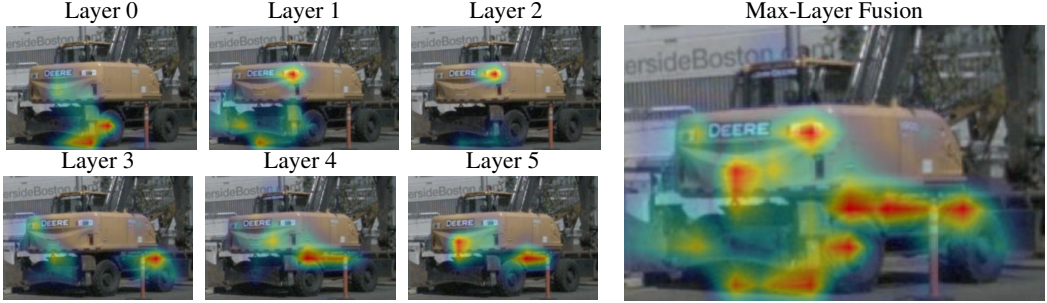

    \begin{minipage}[c]{0.6\linewidth}
        \centering
        \foreach \i in {0,...,5} {
            \begin{subfigure}{0.3\linewidth}
                \centering
                \small{Layer \i}\\
                \includegraphics[width=\linewidth]{layer_\i}
            \end{subfigure}
        }
    \end{minipage}
    \begin{minipage}[c]{0.4\linewidth}
        \centering
        \begin{subfigure}{\linewidth}
            \centering
            \small{Max-Layer Fusion}\\
            \includegraphics[width=\linewidth]{full}
        \end{subfigure}
    \end{minipage}
    \caption{Raw cross-attention $\mathbb{E}_h( A_{CR} )^+$ for all six layers of the decoder. The image on the right depicts the \emph{max-layer} aggregation, where the maximum across all layers is aggregated.}
    \label{fig:aggreg}
\end{figure}

\begin{figure}[t]
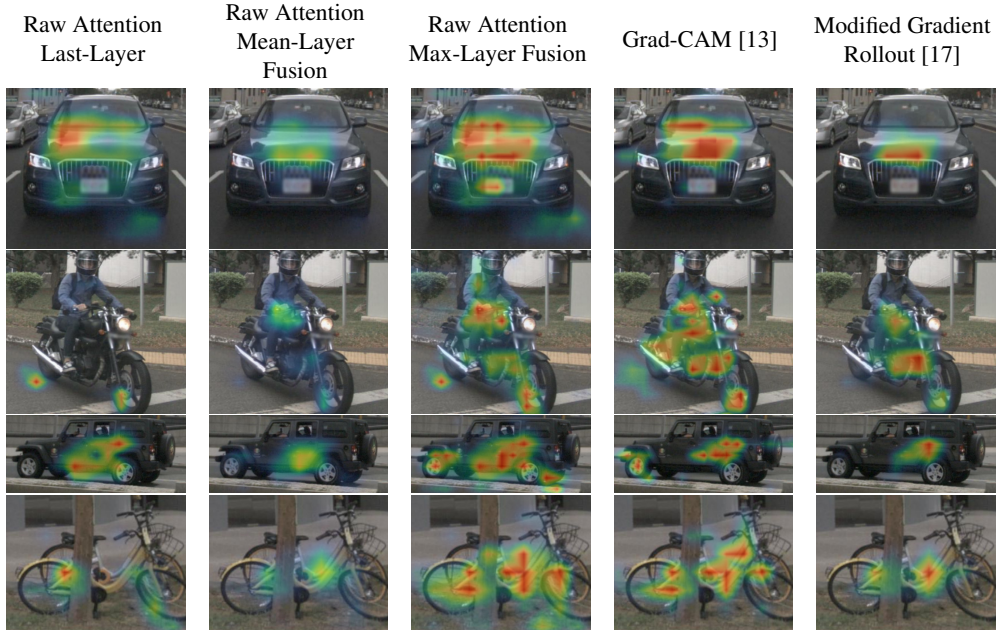

    \centering
    \begin{minipage}{0.17\textwidth}
        \centering
        {\small Raw Attention Last-Layer}
    \end{minipage}
    \hspace{0.01\textwidth}
    \begin{minipage}{0.17\textwidth}
        \centering
        {\small Raw Attention Mean-Layer Fusion}
    \end{minipage}
    \hspace{0.01\textwidth}
    \begin{minipage}{0.17\textwidth}
        \centering
        {\small Raw Attention Max-Layer Fusion}
    \end{minipage}
    \hspace{0.01\textwidth}
    \begin{minipage}{0.17\textwidth}
        \centering
        {\small Grad-CAM \cite{Selvaraju2017}}
    \end{minipage}
    \hspace{0.01\textwidth}
    \begin{minipage}{0.17\textwidth}
        \centering
        {\small Modified Gradient Rollout \cite{Chefer2021_mm}}
    \end{minipage}
    \vspace*{1mm}

    \foreach \i in {0,...,3} { 
      \begin{minipage}{\textwidth}
        \centering
        \begin{minipage}[b]{0.17\textwidth}
          \centering
          \includegraphics[width=\textwidth]{raw_last/\i}
        \end{minipage}
        \hspace{0.01\textwidth}
        \begin{minipage}[b]{0.17\textwidth}
          \centering
          \includegraphics[width=\textwidth]{raw_mean/\i}
        \end{minipage}
        \hspace{0.01\textwidth}
        \begin{minipage}[b]{0.17\textwidth}
          \centering
          \includegraphics[width=\textwidth]{raw_max/\i}
        \end{minipage}
        \hspace{0.01\textwidth}
        \begin{minipage}[b]{0.17\textwidth}
          \centering
          \includegraphics[width=\textwidth]{gradcam/\i}
        \end{minipage}
        \hspace{0.01\textwidth}
        \begin{minipage}[b]{0.17\textwidth}
          \centering
          \includegraphics[width=\textwidth]{gradroll/\i}
        \end{minipage}
      \end{minipage}%
    }
  \caption{Saliency maps generated for different objects classes using all explainability methods.}
  \label{fig:xai_comp}
\end{figure}

We developed a visual exploration tool which let us browse through the nuScenes dataset and investigated the attention mechanism in the model for each head, each layer, different configurations, classes, scenes and aggregation methods. Figure \ref{fig:peds_side} displays raw cross-attention for pedestrians across all six decoder layers. The model pays attention to specific parts of the human body, with a preference for lower torso and legs. Hereby, every layer seems to concentrate on a different part of the body. We could observe this behavior across several different classes, such as \texttt{car}, \texttt{truck} and \texttt{construction vehicle}. Another example is shown in Figure \ref{fig:aggreg} where each layer seems to cover different parts of the construction vehicle. Additional examples are provided in Appendix \ref{appendix:examples}. Furthermore, examples of objects located within the overlapping FOV of two cameras are presented in Appendix \ref{appendix:examples_fov_overlap}. This observation let us to the conclusion, that aggregating the attentions across all decoder layers is immensely important to give a holistic explanation for a certain detection, which resulted in the \emph{mean-layer} and \emph{max-layer} aggregation methods (\emph{cf.} Eq. \ref{eq:mean} and Eq. \ref{eq:max}), which have not been explored previously \cite{Chefer2021_mm}. It seems that the model alternates its attention between specific concepts of a class in every layer to distinguish between other classes, but also find the center, orientation and the dimensions of the OBB. In Figure \ref{fig:xai_comp}, we compare the different explainbility methods on different classes. It becomes clear that if only one layer of the decoder is used, the quality of the saliency map is inferior in many cases as only one "concept" of a class is covered by that layer. The \emph{mean-layer} aggregation gives us a very smooth saliency map for an object, whereas the \emph{max-layer} fusion and \emph{Grad-CAM} give a diversified focus through various parts of the object. Considering \emph{Gradient Rollout}, the resulting saliency map only highlights the most essential areas for detection.

\graphicspath{{results/}} 
\graphicspath{{results/example_aggregated/}}

\graphicspath{{results/}}

\subsection{Perturbation Test}
We use a pre-trained model, iterate over the validation split of the nuScenes dataset and execute the model with the current sample data. With the information of this forward pass, we compute the saliency map for each camera image and we remove a certain percentage of the input: For the \emph{positive perturbation} test, the regions with the highest activation, and for the \emph{negative perturbation}, the regions with the lowest activation are masked, as show in Appendix \ref{appendix:perturbation_test}. The masked regions are filled with the mean of the image. Then, the perturbed camera images are used again as input the to model and the model's output is collected in order to compute the nuScenes Detection Score (NDS) \cite{Caesar2020}. During the test, we gradually increase the percentage of masked input pixels. For the \emph{positive perturbation}, we expect that the NDS decreases quickly as the most important regions are removed first. For the \emph{negative perturbation}, a good saliency generation method would result in a curve with a slow decrease as only irrelevant regions are removed at the beginning of the test. For both tests, we compute the area-under-the-curve (AUC), to measure the impact on the model's performance. To establish a simple baseline for this test, we generate a \emph{random explanation} drawn from a normal distribution $S \sim \mathcal{N}(\mu,\,\sigma^{2})$ for each sample. As shown in Figure \ref{fig:positive_perturbation} and \ref{fig:negative_perturbation}, all five methods perform reasonably well and outperform a \emph{random explanation}. The methods \emph{max-layer} fusion and \emph{Grad-CAM} outperform the other methods, followed by \emph{mean-layer} fusion. As expected, taking only the \emph{last-layer} of the decoder's attentions results in an inferior performance compared to methods that aggregate the attention over all layer. \emph{Gradient Rollout} \cite{Chefer2021_mm} was also less effective than the other methods. The formulation may not work well with the decoder-only transformer architecture, which was not originally designed for it.

\subsection{Saliency Map Sanity Check}
We follow \cite{Adebayo2018} and perform a model parameter randomization test. We randomly initialize an untrained model and generate saliency maps with \emph{max-layer} fusion. The test determines if the saliency maps depend on the learned parameters of the model. Some model and saliency generation methods are insensitive to the properties of the trained parameters, if the saliency maps are insensitive to the learned parameters, then the saliency maps of the trained and randomly initialized model will be similar \cite{Adebayo2018}. We could observe that the randomly initialized model does not produce reasonable saliency maps. Examples are shown in Appendix \ref{appendix:sanity_check}. Hence, our approach passes the sanity check.


\begin{figure}
\begin{tikzpicture}
\pgfplotstableread{data/pp_random.dat}{\random}
\pgfplotstableread{data/pp_raw_max.dat}{\rawmax}
\pgfplotstableread{data/pp_raw_mean.dat}{\rawmean}
\pgfplotstableread{data/pp_raw_last.dat}{\rawlast}
\pgfplotstableread{data/pp_grad_cam.dat}{\gradcam}
\pgfplotstableread{data/pp_gradient_rollout.dat}{\gradientrollout}
\begin{axis}[
    xmin = 0, xmax = 100,
    ymin = 0.02, ymax = 0.40,
    xtick distance = 5,
    ytick distance = 0.1,
    grid = both,
    minor tick num = 1,
    major grid style = {lightgray},
    minor grid style = {lightgray!25},
    width = 1.0\linewidth,
    height = 0.40\linewidth,
    legend cell align = {left},
    legend pos = north east,
    legend style={font=\scriptsize},
    xlabel={Percentage of Masked Input Pixels [\%]},
    ylabel={nuScenes Detection Score [NDS]},
]
\addplot[blue,  mark=*] table [x = {perc}, y = {nds}] {\random};
\addplot[red,   mark=square*] table [x ={perc}, y = {nds}] {\rawmax};
\addplot[gray,  mark=diamond*] table [x ={perc}, y = {nds}] {\rawmean};
\addplot[cyan,  mark=oplus] table [x ={perc}, y = {nds}] {\rawlast};
\addplot[teal,  mark=x] table [x = {perc}, y = {nds}] {\gradcam};
\addplot[green, mark=triangle*] table [x = {perc}, y = {nds}] {\gradientrollout};
\legend{
    Random Explanation - AUC 18.38,
    Attention Max-Layer Fusion - \textbf{AUC 8.92},
    Attention Mean-Layer Fusion - AUC 11.35,
    Attention Last-Layer - AUC 11.17,
    Grad-CAM - AUC 9.86,
    Gradient Rollout - AUC 11.09
}
\end{axis}
\end{tikzpicture}
\caption{Positive perturbation test evaluated on the nuScenes validation set. Smaller AUC is better.}
\end{figure}
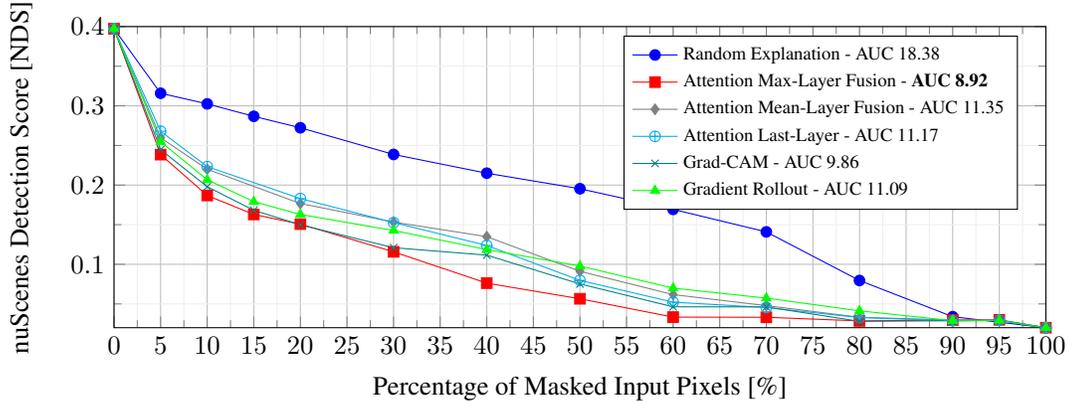
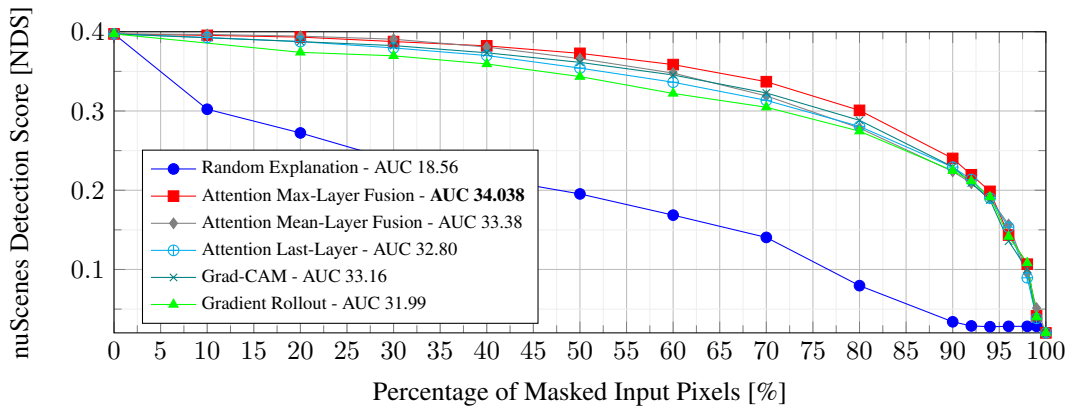
\begin{figure}
\begin{tikzpicture}
\pgfplotstableread{data/np_random.dat}{\random}
\pgfplotstableread{data/np_raw_max.dat}{\rawmax}
\pgfplotstableread{data/np_raw_mean.dat}{\rawmean}
\pgfplotstableread{data/np_raw_last.dat}{\rawlast}
\pgfplotstableread{data/np_grad_cam.dat}{\gradcam}
\pgfplotstableread{data/np_gradient_rollout.dat}{\gradientrollout}
\begin{axis}[
    xmin = 0, xmax = 100,
    ymin = 0.02, ymax = 0.40,
    xtick distance = 5,
    ytick distance = 0.1,
    grid = both,
    minor tick num = 1,
    major grid style = {lightgray},
    minor grid style = {lightgray!25},
    width = 1\linewidth,
    height = 0.40\linewidth,
    legend cell align = {left},
    legend pos = south west,
    legend style={font=\scriptsize},
    xlabel={Percentage of Masked Input Pixels [\%]},
    ylabel={nuScenes Detection Score [NDS]},
]
\addplot[blue,  mark=*] table [x = {perc}, y = {nds}] {\random};
\addplot[red,   mark=square*] table [x ={perc}, y = {nds}] {\rawmax};
\addplot[gray,  mark=diamond*] table [x ={perc}, y = {nds}] {\rawmean};
\addplot[cyan,  mark=oplus] table [x ={perc}, y = {nds}] {\rawlast};
\addplot[teal,  mark=x] table [x = {perc}, y = {nds}] {\gradcam};
\addplot[green, mark=triangle*] table [x = {perc}, y = {nds}] {\gradientrollout};
\legend{
    Random Explanation - AUC 18.56,
    Attention Max-Layer Fusion - \textbf{AUC 34.038},
    Attention Mean-Layer Fusion - AUC 33.38,
    Attention Last-Layer - AUC 32.80,
    Grad-CAM - AUC 33.16,
    Gradient Rollout - AUC 31.99
}
\end{axis}
\end{tikzpicture}
\caption{Negative perturbation test evaluated on the nuScenes validation set. Larger AUC is better.}
\end{figure}

\section{Conclusion}
We introduced an approach for generating saliency maps using a transformer-based multi-camera model in the context of 3D object detection. Our goal was to provide insights into which regions of input images are most influential in determining object detection results. Our approach, based on raw cross-attention, proved to be more efficient than gradient-based methods. Our perturbation test demonstrated the effectiveness of our approach. Visual exploration of attention maps showed that the transformer's attention alternates between different concepts of a class in different layers to distinguish between classes and to determine object properties. Aggregating attention across different layers was crucial for a comprehensive explanation.

The current method is restricted to generating saliency maps solely for queries exceeding the detection threshold. Queries that produce detections with low classification scores are not considered. Additionally, the resolution of the generated saliency maps is relatively coarse, enabling the highlighting of object concepts but not fine details. In the future, we want to extend the approach to a multi-modal transformer model which utilizes LiDAR and multi-view camera images as input. We aim to enhance the transparency of such end-to-end model by tracing the origins of detection, attributing them to the specific sensors that contributed to their identification.

\begin{ack}
This work has received funding from the European Union's Horizon Europe Research and Innovation Programme under Grant Agreement No 101076754 - AIthena project.
\end{ack}

\bibliographystyle{unsrt}
\bibliography{neurips_2023_ml4ad}

\newpage
\appendix
\section{Implementation Details}\label{appendix:algorithms}
\begin{minipage}[t]{0.48\textwidth}
\begin{algorithm}[H]
\scriptsize
\SetKwInOut{Input}{Input}
\SetKwInOut{Output}{Output}

\Input{$A_{CR} \in \mathbb{R}^{n_{c} \times n_{l} \times n_{h} \times n_{q} \times n_{t}}$}
\Output{$S \in \mathbb{R}^{n_{c} \times H \times W}$}

$S \gets \emptyset$ \\


\For {$c=1$ to $n_{c}$} {
\# select camera $c$ \\
$S^c \gets \emptyset$ \\
$A_{CR}^{c} \in \mathbb{R}^{n_l \times n_h \times n_q \times n_t} \gets A_{CR} $ \\
$A_{CR}^{ch} \in \mathbb{R}^{n_l \times n_q \times n_t}  \gets \mbox{max}_{h \in \{1,\dots, n_{h}\} } A_{CR}^{c}$ \\
$A_{CR}^{clh} \in \mathbb{R}^{n_q \times n_t}  \gets \mbox{max}_{l \in \{1,\dots, n_{l}\} } A_{CR}^{ch}$

\For {$q=1$ to $n_{q}$} {

\If{ $y_q$ > theshold}{

$S^{q} \in \mathbb{R}^{H'\times W'} \gets A_{CR}^{lcq} \in \mathbb{R}^{n_t}$ \\
$S^{q} \in \mathbb{R}^{H \times W} \gets S^{q} \in \mathbb{R}^{H' \times W'}$ \\
$S^c \in \mathbb{R}^{H \times W} \gets S^c + S^{q} \in \mathbb{R}^{H \times W}$

} 
} 
$S \gets S^c$ \\
} 
\caption{Raw Attention Max-Layer}
\label{alg:raw_attention_max_layer_fusion}
\end{algorithm}
\end{minipage}
\hfill
\begin{minipage}[t]{0.48\textwidth}
\begin{algorithm}[H]
\scriptsize
\SetKwInOut{Input}{Input}
\SetKwInOut{Output}{Output}

\Input{$A_{CR} \in \mathbb{R}^{n_{c} \times n_{l} \times n_{h} \times n_{q} \times n_{t}}$}
\Output{$S \in \mathbb{R}^{n_{c} \times H \times W}$}

$S \gets \emptyset$ \\
\# compute gradient w.t.r. object scores \\
$\nabla A_{CR} \in \mathbb{R}^{n_{c} \times n_{l} \times n_{h} \times n_{q} \times n_{t}} \gets \frac{\partial y}{\partial A}$ \\


\For {$c=1$ to $n_{c}$} {
\# select camera $c$ \\
$S^c \gets \emptyset$ \\
$ A_{CR}^{c} \in \mathbb{R}^{n_l \times n_h \times n_q \times n_t} \gets A_{CR} $ \\
$\nabla A_{CR}^{c} \in \mathbb{R}^{n_l \times n_h \times n_q \times n_t} \gets \nabla A_{CR} $ \\
$A_{CR}^{ch} \in \mathbb{R}^{n_l \times n_q \times n_t} \gets \mathbb{E}_h\left(({\nabla A_{CR}^{c}} \cdot A_{CR}^{c})^+\right)$ \\
$A_{CR}^{clh} \in \mathbb{R}^{n_q \times n_t}  \gets \mathbb{E}_l\left( A_{CR}^{ch}\right)$

\For {$q=1$ to $n_{q}$} {

\If{ $y_q$ > theshold}{

$S^{q} \in \mathbb{R}^{H'\times W'} \gets A_{CR}^{clhq} \in \mathbb{R}^{n_t}$ \\
$S^{q} \in \mathbb{R}^{H \times W} \gets S^{q} \in \mathbb{R}^{H' \times W'}$ \\
$S^c \in \mathbb{R}^{H \times W} \gets S^c + S^{q} \in \mathbb{R}^{H \times W}$

} 
} 
$S \gets S^c$ \\
} 
\caption{Grad-CAM}
\label{alg:grad_cam}
\end{algorithm}

\end{minipage}
\begin{figure}[ht]
    \include{algo/gradient_rollout}
\end{figure}


\newpage
\section{Perturbation Test}\label{appendix:perturbation_test}
A perturbation tests is a popular approach to evaluate the effectiveness of explainability methods \cite{Chefer2021, Chefer2021_mm}. We designed the following perturbation test for our 3D object detection setup: We use the whole nuScenes validation split and compute the saliency maps for each camera in each sample. Then we mask a certain percentage of the image input pixels based on the saliency maps and use this perturbed images to compute the nuScenes detection score (NDS). For the \emph{positive} perturbation test we gradually mask the areas in the input camera images with the highest saliency activation, as shown in Fig. \ref{fig:positive_perturbation}. For the \emph{negative} perturbation test the area with the lowest activation is masked first, visualized by Fig. \ref{fig:negative_perturbation}.
\def\indices{{0,5,10,15}}
\graphicspath{{img/pos_perturbation/}}
\begin{figure}[!htbp]
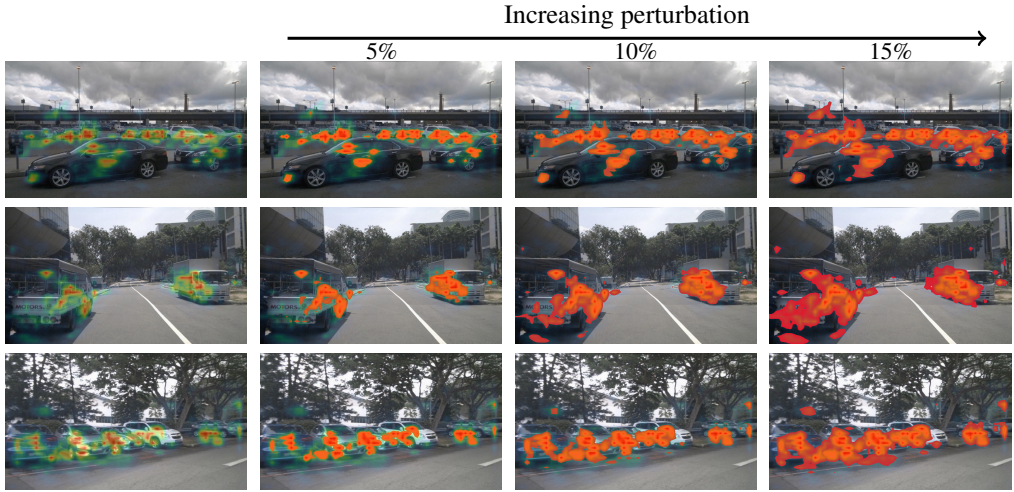

    \begin{minipage}{0.95\textwidth}
        \raggedleft
        \begin{tikzpicture}
            \draw [->, line width=0.4mm] (0.3\textwidth,0) -- (\textwidth,0);
            \node [above] at (0.64\textwidth,0) {Increasing perturbation};
        \end{tikzpicture}
    \end{minipage}

    \begin{minipage}{\textwidth}
        \centering
        \foreach \i in {0,...,3} {
            \begin{subfigure}[b]{0.23\textwidth}
                \centering
                \pgfmathsetmacro{\im}{\indices[\i]}
                \ifnum\i>0
                   {\small \pgfmathparse{\indices[\i]}\pgfmathresult \%}\\
                \fi
                \includegraphics[width=\textwidth]{1462/\im}
            \end{subfigure}
        }
        \vspace*{1mm}

        \foreach \i in {0,...,3} {  
            \begin{subfigure}[b]{0.23\textwidth}
                \centering
                \pgfmathsetmacro{\im}{\indices[\i]}
                \includegraphics[width=\textwidth]{4260/\im}
            \end{subfigure}
        }
        \vspace*{1mm}

        \foreach \i in {0,...,3} {  
            \begin{subfigure}[b]{0.23\textwidth}
                \centering
                \pgfmathsetmacro{\im}{\indices[\i]}
                \includegraphics[width=\textwidth]{4826/\im}
            \end{subfigure}
        }
        \vspace*{1mm}

        \caption{Visualization of the \emph{positive perturbation} test on three sample images from the nuScenes dataset while increasing the number of masked pixels. Masked pixels, here shown in red, are filled with the mean of the image.}
        \label{fig:positive_perturbation}
    \end{minipage}
\end{figure}

\def\indices{{0,70,80,90}}
\graphicspath{{img/neg_perturbation/}}
\begin{figure}[!htbp]
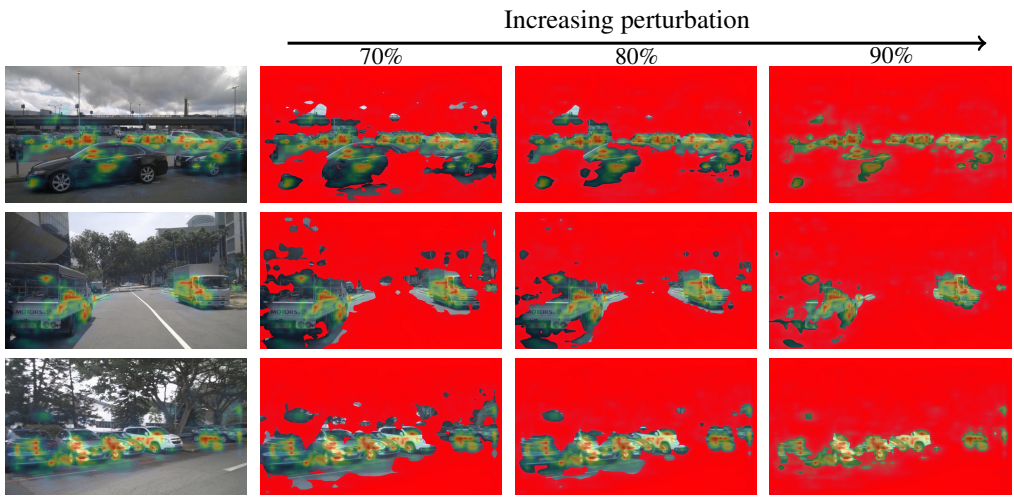

    \begin{minipage}{0.95\textwidth}
        \raggedleft
        \begin{tikzpicture}
            \draw [->, line width=0.4mm] (0.3\textwidth,0) -- (\textwidth,0);
            \node [above] at (0.64\textwidth,0) {Increasing perturbation};
        \end{tikzpicture}
    \end{minipage}

    \begin{minipage}{\textwidth}
        \centering
        \foreach \i in {0,...,3} { 
            \begin{subfigure}[b]{0.23\textwidth}
                \centering
                \pgfmathsetmacro{\im}{\indices[\i]}
                \ifnum\i>0
                   {\small \pgfmathparse{\indices[\i]}\pgfmathresult \%}\\
                \fi
                \includegraphics[width=\textwidth]{1462/\im}
            \end{subfigure}
        }
        \vspace*{1mm}

        \foreach \i in {0,...,3} {  
            \begin{subfigure}[b]{0.23\textwidth}
                \centering
                \pgfmathsetmacro{\im}{\indices[\i]}
                \includegraphics[width=\textwidth]{4260/\im}
            \end{subfigure}
        }
        \vspace*{1mm}

        \foreach \i in {0,...,3} {  
            \begin{subfigure}[b]{0.23\textwidth}
                \centering
                \pgfmathsetmacro{\im}{\indices[\i]}
                \includegraphics[width=\textwidth]{4826/\im}
            \end{subfigure}
        }
        \vspace*{1mm}

        \caption{Visualization of the \emph{negative perturbation} test on three sample images from the nuScenes dataset while increasing the number of masked pixels. Masked pixels, here shown in red, are filled with the mean of the image.}
        \label{fig:negative_perturbation}
    \end{minipage}
\end{figure}

\newpage
\section{Sanity Checks for Saliency Maps}\label{appendix:sanity_check}
\begin{figure}[!htbp]
  \centering
  \includegraphics[width=1\textwidth]{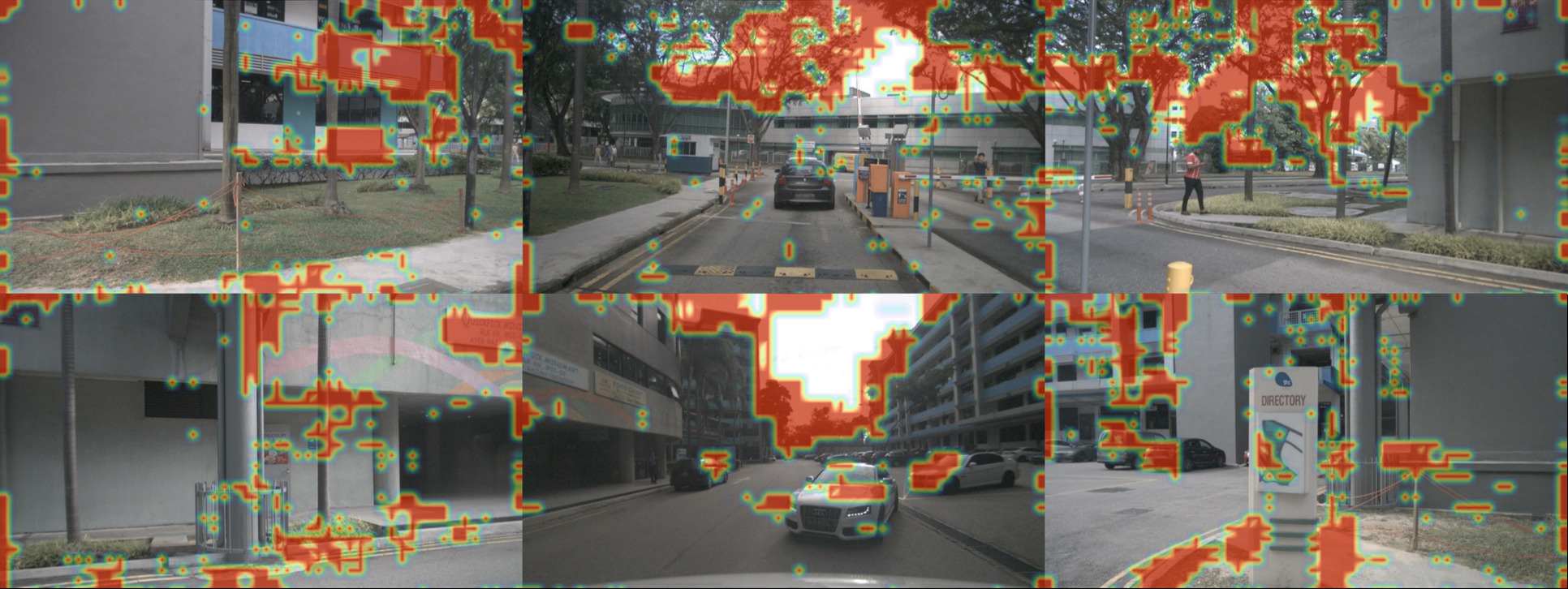}
  \caption{Raw saliency map when transformer decoder and backbone weights are randomly initialized. The resulting saliency maps are meaningless and not comparable to the saliency maps of the trained model. This indicates that the sanity check for saliency maps has been passed.}
  \label{fig:sanity_check_1}
\end{figure}

\begin{figure}[!htbp]
  \centering
  \includegraphics[width=1\textwidth]{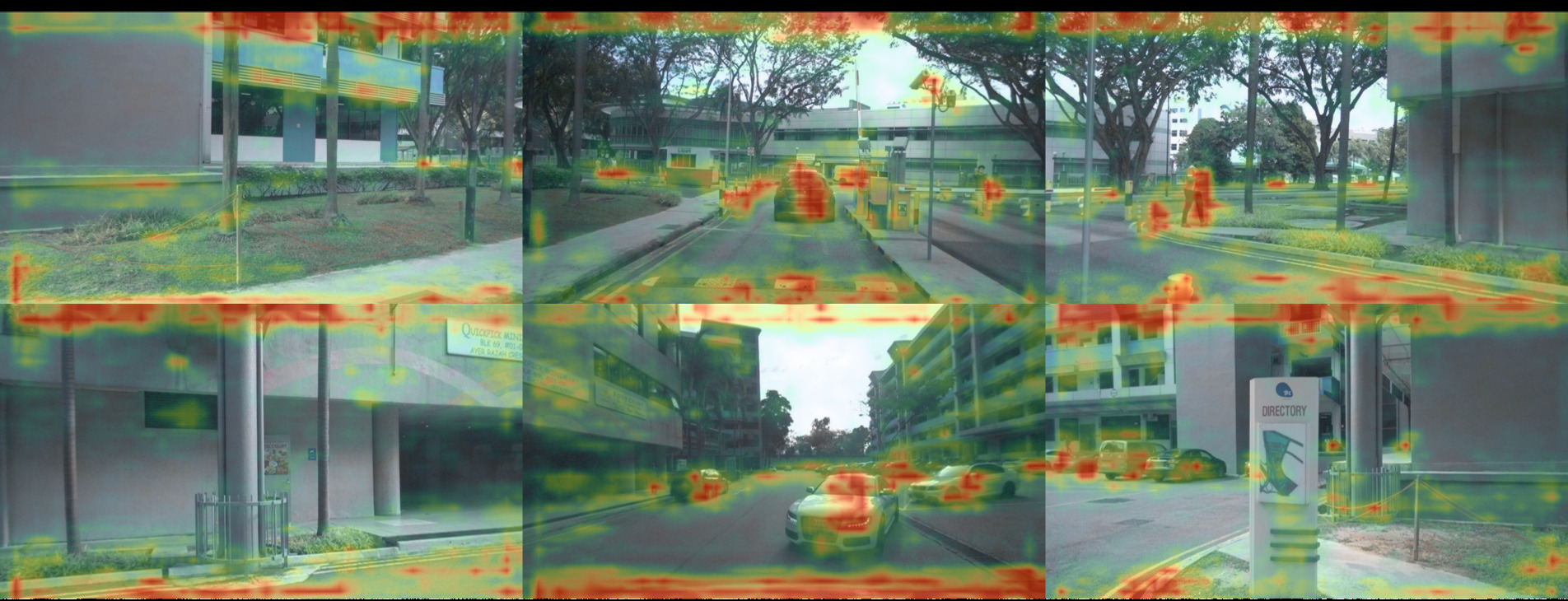}
  \caption{Raw saliency map output when the ResNet backbone weights are initialized from an object detection checkpoint, but the weights of the transformer decoder are randomly initialized. This settings seems to highlight some areas of relevant objects, but also a lot of noise is generated. This underlines that the saliency maps generated with our approach are in fact sensitive to the properties of the transformer model weights.}
  \label{fig:sanity_check_2}
\end{figure}

We follow \cite{Adebayo2018} and conduct a \emph{model parameter randomization test} in which we generate saliency maps of a randomly initialized untrained network, as shown in Fig. \ref{fig:sanity_check_1}. The generated saliency maps differ substantially from the saliency maps of the trained model, which indicates that the sanity check has been passed. We perform another test, by initializing the ResNet backbone from a checkpoint while keeping the transfomer weights uninitialized. This test should show that the saliency maps do not dependent solely on the CNN-based backbone, but rather from the interplay between backbone and transformer. As shown in Fig. \ref{fig:sanity_check_2}, the obtained saliency maps are very noisy, but we can also observe some highlights on relevant objects, leading to the conclusion that the saliency maps in our approach are in fact sensitive to the properties of the transformer decoder.
\graphicspath{{img/sanity_check}} 

\newpage
\section{Examples}\label{appendix:examples}

\graphicspath{{examples_techniques/}}
\begin{figure}[!htbp]
    \centering
    \begin{minipage}{0.17\textwidth}
        \centering
        {\small Raw Attention Last-Layer}
    \end{minipage}
    \hspace{0.01\textwidth}
    \begin{minipage}{0.17\textwidth}
        \centering
        {\small Raw Attention Mean-Layer Fusion}
    \end{minipage}
    \hspace{0.01\textwidth}
    \begin{minipage}{0.17\textwidth}
        \centering
        {\small Raw Attention Max-Layer Fusion}
    \end{minipage}
    \hspace{0.01\textwidth}
    \begin{minipage}{0.17\textwidth}
        \centering
        {\small Grad-CAM \cite{Selvaraju2017}}
    \end{minipage}
    \hspace{0.01\textwidth}
    \begin{minipage}{0.17\textwidth}
        \centering
        {\small Modified Gradient Rollout \cite{Chefer2021_mm}}
    \end{minipage}
    \vspace*{1mm}

    \foreach \i in {0,...,6} { 
      \begin{minipage}{\textwidth}
        \centering
        \begin{minipage}[b]{0.17\textwidth}
          \centering
          \includegraphics[width=\textwidth]{raw_last/\i}
        \end{minipage}
        \hspace{0.01\textwidth}
        \begin{minipage}[b]{0.17\textwidth}
          \centering
          \includegraphics[width=\textwidth]{raw_mean/\i}
        \end{minipage}
        \hspace{0.01\textwidth}
        \begin{minipage}[b]{0.17\textwidth}
          \centering
          \includegraphics[width=\textwidth]{raw_max/\i}
        \end{minipage}
        \hspace{0.01\textwidth}
        \begin{minipage}[b]{0.17\textwidth}
          \centering
          \includegraphics[width=\textwidth]{gradcam/\i}
        \end{minipage}
        \hspace{0.01\textwidth}
        \begin{minipage}[b]{0.17\textwidth}
          \centering
          \includegraphics[width=\textwidth]{gradroll/\i}
        \end{minipage}
      \end{minipage}
    }
    \vspace*{1mm}
  \caption{Saliency maps generated for Raw Attention (Last, Mean, Max), Grad-CAM and Gradient Rollout for different objects.}
  \label{fig:xai_examples1}
\end{figure}

\begin{figure}[!htbp]
    \centering
    \begin{minipage}{0.17\textwidth}
        \centering
        {\small Raw Attention Last-Layer}
    \end{minipage}
    \hspace{0.01\textwidth}
    \begin{minipage}{0.17\textwidth}
        \centering
        {\small Raw Attention Mean-Layer Fusion}
    \end{minipage}
    \hspace{0.01\textwidth}
    \begin{minipage}{0.17\textwidth}
        \centering
        {\small Raw Attention Max-Layer Fusion}
    \end{minipage}
    \hspace{0.01\textwidth}
    \begin{minipage}{0.17\textwidth}
        \centering
        {\small Grad-CAM \cite{Selvaraju2017}}
    \end{minipage}
    \hspace{0.01\textwidth}
    \begin{minipage}{0.17\textwidth}
        \centering
        {\small Modified Gradient Rollout \cite{Chefer2021_mm}}
    \end{minipage}
    \vspace*{1mm}

    \foreach \i in {6,...,12} { 
      \begin{minipage}{\textwidth}
        \centering
        \begin{minipage}[b]{0.17\textwidth}
          \centering
          \includegraphics[width=\textwidth]{raw_last/\i}
        \end{minipage}
        \hspace{0.01\textwidth}
        \begin{minipage}[b]{0.17\textwidth}
          \centering
          \includegraphics[width=\textwidth]{raw_mean/\i}
        \end{minipage}
        \hspace{0.01\textwidth}
        \begin{minipage}[b]{0.17\textwidth}
          \centering
          \includegraphics[width=\textwidth]{raw_max/\i}
        \end{minipage}
        \hspace{0.01\textwidth}
        \begin{minipage}[b]{0.17\textwidth}
          \centering
          \includegraphics[width=\textwidth]{gradcam/\i}
        \end{minipage}
        \hspace{0.01\textwidth}
        \begin{minipage}[b]{0.17\textwidth}
          \centering
          \includegraphics[width=\textwidth]{gradroll/\i}
        \end{minipage}
      \end{minipage}
    }
    \vspace*{1mm}
  \caption{Saliency maps generated for Raw Attention (Last, Mean, Max), Grad-CAM and Gradient Rollout for different objects.}
  \label{fig:xai_examples2}
\end{figure}

\begin{figure}[!htb]
    \centering
    \begin{minipage}{0.17\textwidth}
        \centering
        {\small Raw Attention Last-Layer}
    \end{minipage}
    \hspace{0.01\textwidth}
    \begin{minipage}{0.17\textwidth}
        \centering
        {\small Raw Attention Mean-Layer Fusion}
    \end{minipage}
    \hspace{0.01\textwidth}
    \begin{minipage}{0.17\textwidth}
        \centering
        {\small Raw Attention Max-Layer Fusion}
    \end{minipage}
    \hspace{0.01\textwidth}
    \begin{minipage}{0.17\textwidth}
        \centering
        {\small Grad-CAM \cite{Selvaraju2017}}
    \end{minipage}
    \hspace{0.01\textwidth}
    \begin{minipage}{0.17\textwidth}
        \centering
        {\small Modified Gradient Rollout \cite{Chefer2021_mm}}
    \end{minipage}
    \vspace*{1mm}

    \foreach \i in {12,...,19} { 
      \begin{minipage}{\textwidth}
        \centering
        \begin{minipage}[b]{0.17\textwidth}
          \centering
          \includegraphics[width=\textwidth]{raw_last/\i}
        \end{minipage}
        \hspace{0.01\textwidth}
        \begin{minipage}[b]{0.17\textwidth}
          \centering
          \includegraphics[width=\textwidth]{raw_mean/\i}
        \end{minipage}
        \hspace{0.01\textwidth}
        \begin{minipage}[b]{0.17\textwidth}
          \centering
          \includegraphics[width=\textwidth]{raw_max/\i}
        \end{minipage}
        \hspace{0.01\textwidth}
        \begin{minipage}[b]{0.17\textwidth}
          \centering
          \includegraphics[width=\textwidth]{gradcam/\i}
        \end{minipage}
        \hspace{0.01\textwidth}
        \begin{minipage}[b]{0.17\textwidth}
          \centering
          \includegraphics[width=\textwidth]{gradroll/\i}
        \end{minipage}
      \end{minipage}
    }
    \vspace*{1mm}
  \caption{Saliency maps generated for Raw Attention (Last, Mean, Max), Grad-CAM and Gradient Rollout for different objects.}
  \label{fig:xai_examples3}
\end{figure}

\graphicspath{{examples_layers/car}} 
\begin{figure}[!htb]
    \foreach \example in {0,...,6} {
        \centering
        \ifnum\example=0
            \foreach \i in {0,...,5} {  
                \begin{minipage}{0.15\textwidth}
                    \centering
                    Layer \i  
                \end{minipage}
            }
        \fi
        \foreach \i in {0,...,5} {  
            \begin{subfigure}[b]{0.15\textwidth}
                \includegraphics[width=1\textwidth]{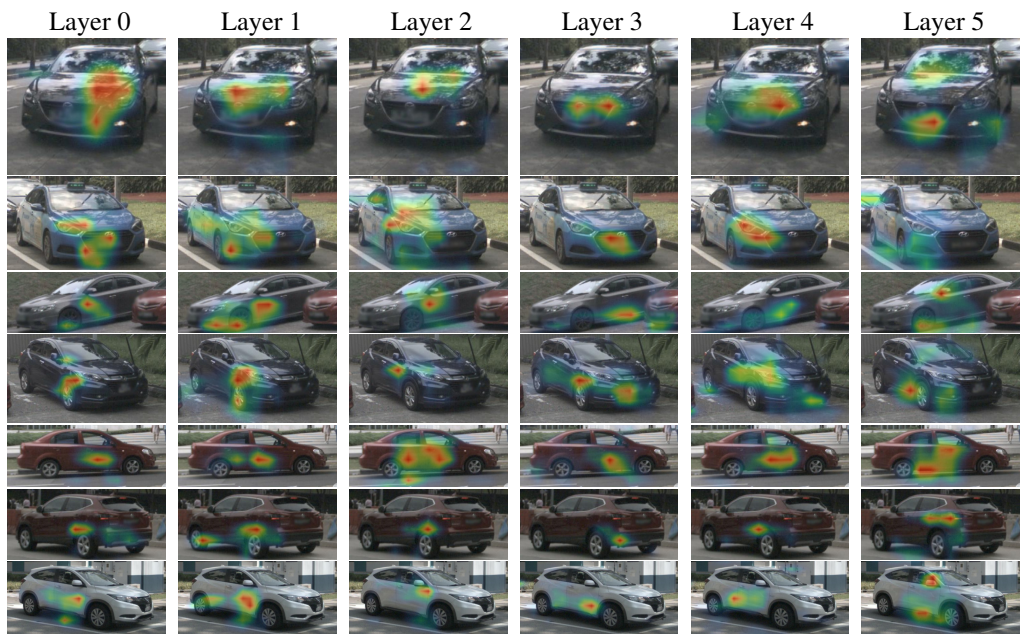}
            \end{subfigure}
        }
        
    }
\label{fig:ex_car}
\caption{Raw cross-attention $\mathbb{E}_h( A_{CR} )^+$ examples for class \texttt{car}.}
\end{figure}

\graphicspath{{examples_layers/bicycle}} 
\begin{figure}[!htb]
    \foreach \example in {0,1,2,3,5} {
        \centering
        \ifnum\example=0
            \foreach \i in {0,...,5} {
                \begin{minipage}{0.15\textwidth}
                    \centering
                    Layer \i  
                \end{minipage}
            }
        \fi
        \foreach \i in {0,...,5} {
            \begin{subfigure}[b]{0.15\textwidth}
                \includegraphics[width=1\textwidth]{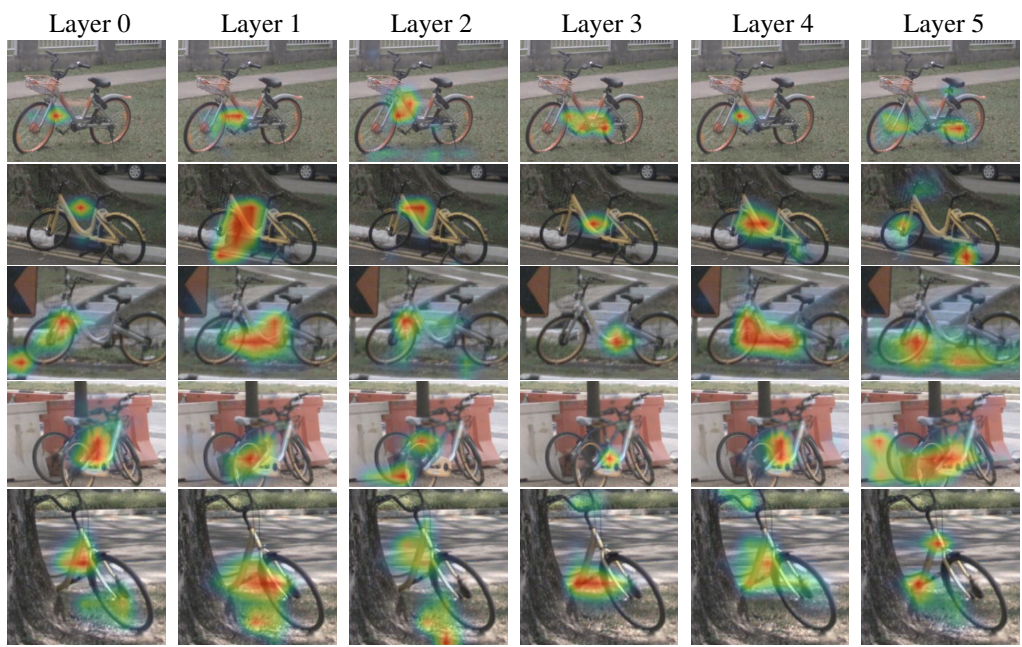}
            \end{subfigure}
        }
        
    }
\label{fig:ex_bicycle}
\caption{Raw cross-attention $\mathbb{E}_h( A_{CR} )^+$ examples for class \texttt{bicycles}.}
\end{figure}

\graphicspath{{examples_layers/truck}} 
\begin{figure}[!htb]
    \foreach \example in {0,...,6} {
        \centering
        \ifnum\example=0
            \foreach \i in {0,...,5} {  
                \begin{minipage}{0.15\textwidth}
                    \centering
                    Layer \i  
                \end{minipage}
            }
        \fi
        \foreach \i in {0,...,5} {  
            \begin{subfigure}[b]{0.15\textwidth}
                \includegraphics[width=1\textwidth]{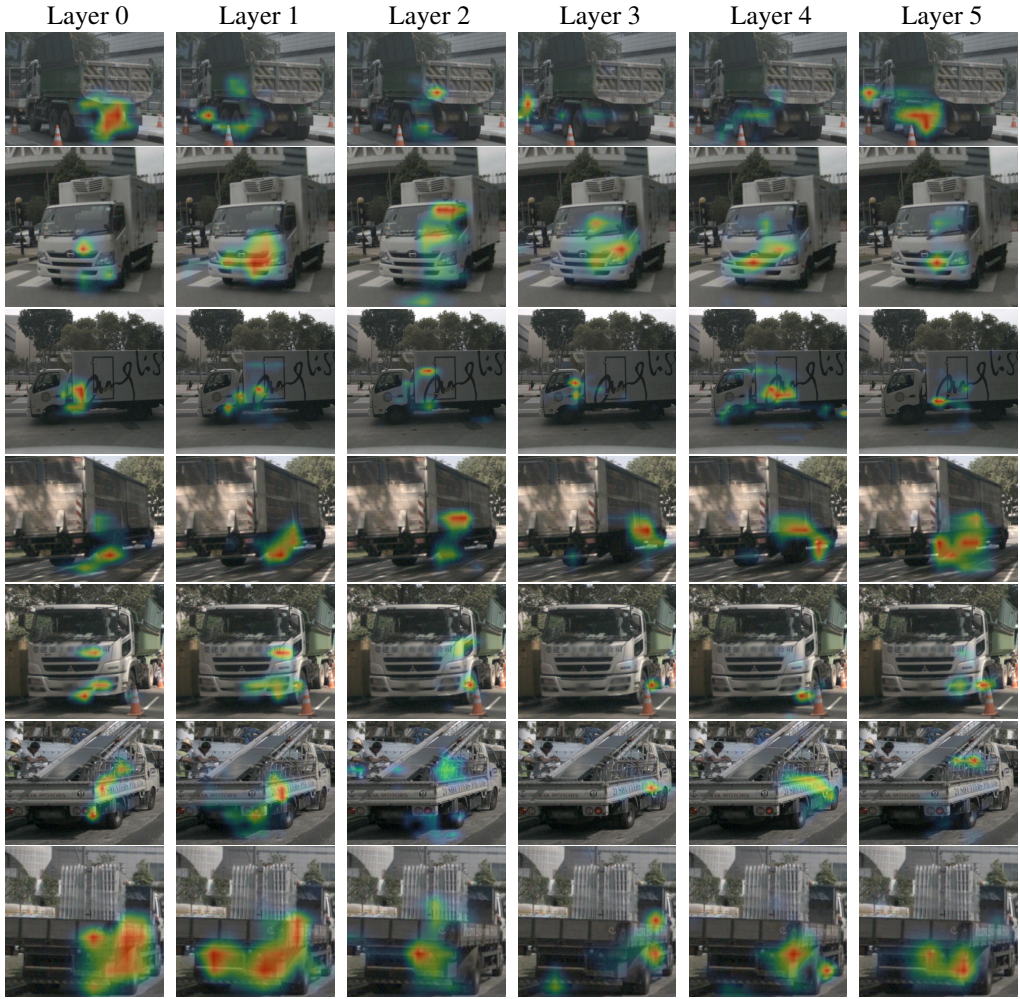}
            \end{subfigure}
        }
        
    }
\label{fig:ex_trucks}
\caption{Raw cross-attention $\mathbb{E}_h( A_{CR} )^+$ examples for class \texttt{trucks}.}
\end{figure}

\graphicspath{{examples_layers/bus}} 
\begin{figure}[!htb]
    \foreach \example in {0,...,4} {
        \centering
        \ifnum\example=0
            \foreach \i in {0,...,5} {  
                \begin{minipage}{0.15\textwidth}
                    \centering
                    Layer \i  
                \end{minipage}
            }
        \fi
        \foreach \i in {0,...,5} {  
            \begin{subfigure}[b]{0.15\textwidth}
                \includegraphics[width=1\textwidth]{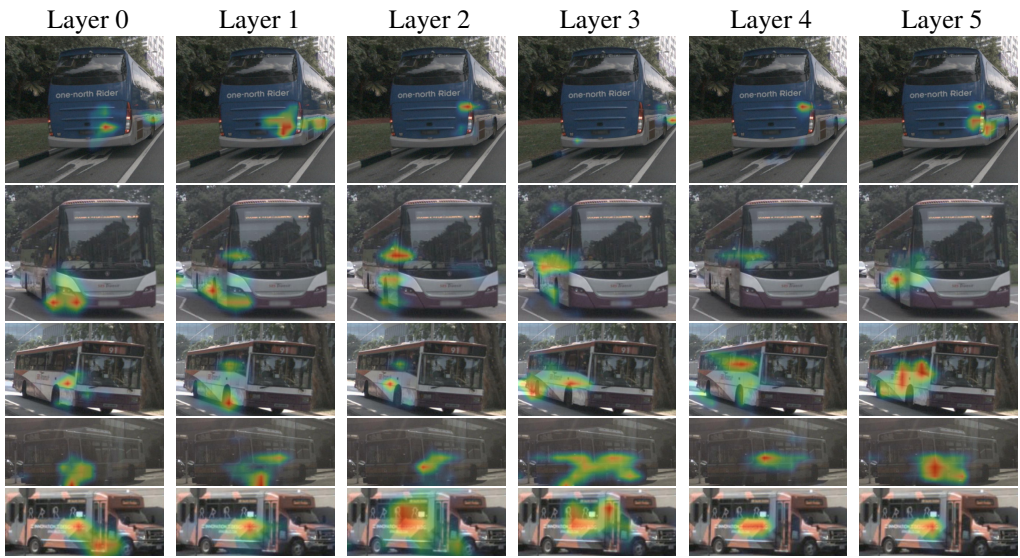}
            \end{subfigure}
        }
        
    }
\label{fig:ex_buses}
\caption{Raw cross-attention $\mathbb{E}_h( A_{CR} )^+$ examples for class \texttt{buses}.}
\end{figure}

\graphicspath{{examples_layers/pedestrian}} 
\begin{figure}[!htb]
    \foreach \example in {0,...,3} {
        \centering
        \ifnum\example=0
            \foreach \i in {0,...,5} {  
                \begin{minipage}{0.14\textwidth}
                    \centering
                    Layer \i  
                \end{minipage}
            }
        \fi
        \foreach \i in {0,...,5} {  
            \begin{subfigure}[b]{0.14\textwidth}
                \includegraphics[width=1\textwidth]{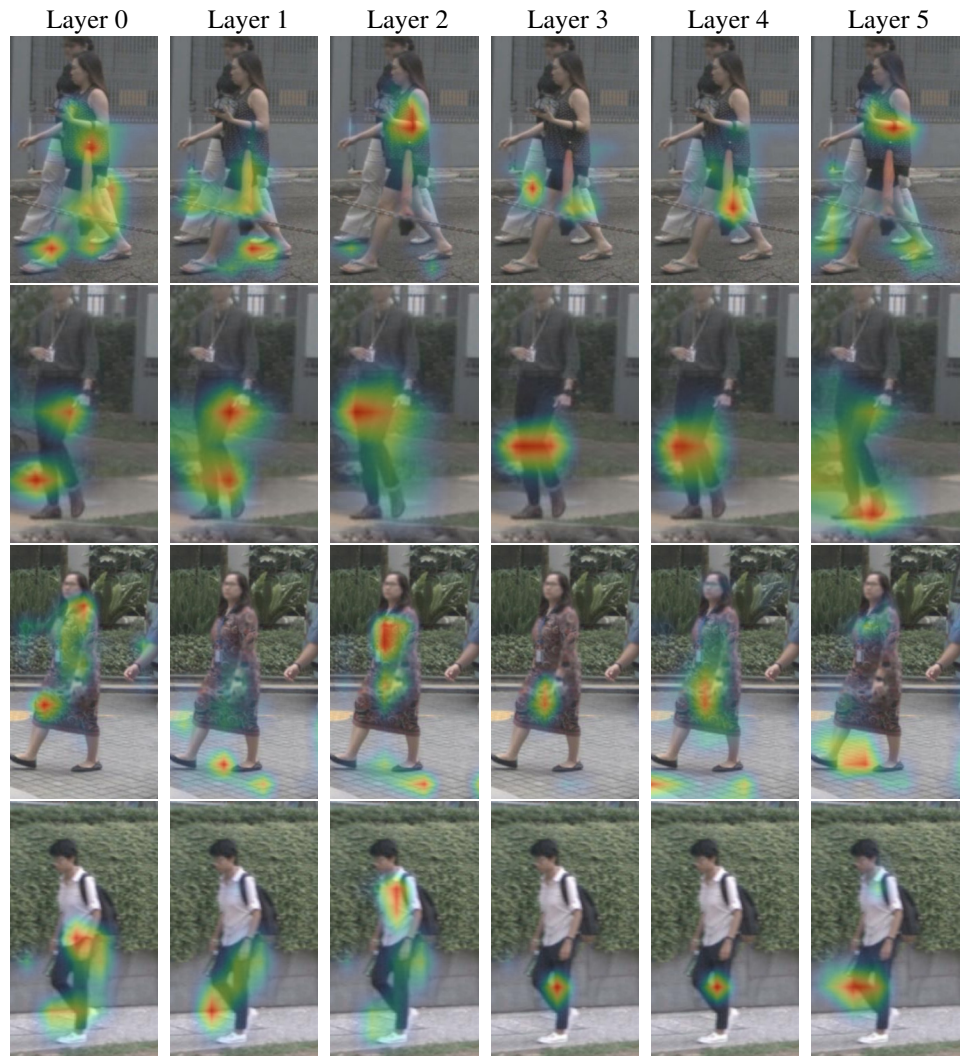}
            \end{subfigure}
        }
        
    }
\label{fig:ex_pedestrians}
\caption{Raw cross-attention $\mathbb{E}_h( A_{CR} )^+$ examples for class \texttt{pedestrians}.}
\end{figure}

\graphicspath{{examples_layers/motorcycle}} 
\begin{figure}[!htb]
    \foreach \example in {0,...,6} {
        \centering
        \ifnum\example=0
            \foreach \i in {0,...,5} {  
                \begin{minipage}{0.15\textwidth}
                    \centering
                    Layer \i  
                \end{minipage}
            }
        \fi
        \foreach \i in {0,...,5} {  
            \begin{subfigure}[b]{0.15\textwidth}
                \includegraphics[width=1\textwidth]{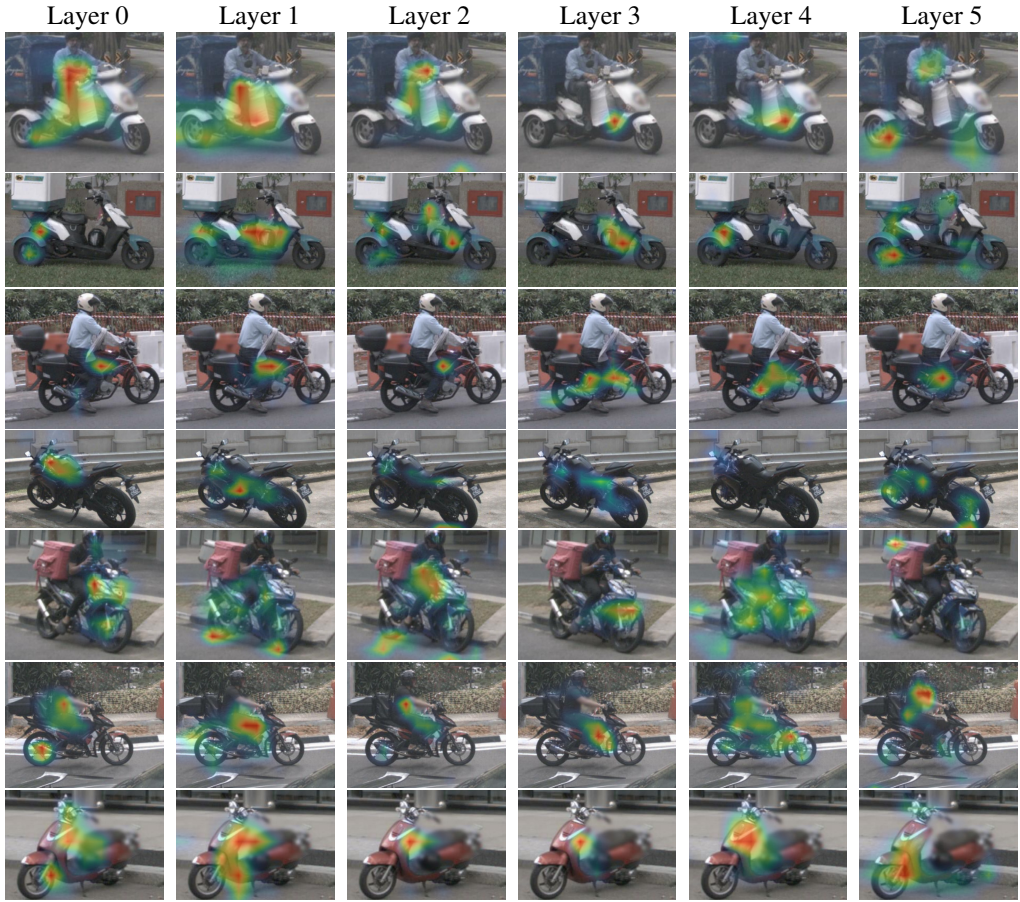}
            \end{subfigure}
        }
        
    }
\label{fig:ex_motorcycle}
\caption{Raw cross-attention $\mathbb{E}_h( A_{CR} )^+$ examples for class \texttt{motorcycles}.}
\end{figure}

\graphicspath{{examples_layers/construction_vehicle}} 
\begin{figure}[!htb]
    \foreach \example in {0,...,4} {
        \centering
        \ifnum\example=0
            \foreach \i in {0,...,5} {  
                \begin{minipage}{0.15\textwidth}
                    \centering
                    Layer \i  
                \end{minipage}
            }
        \fi
        \foreach \i in {0,...,5} {  
            \begin{subfigure}[b]{0.15\textwidth}
                \includegraphics[width=1\textwidth]{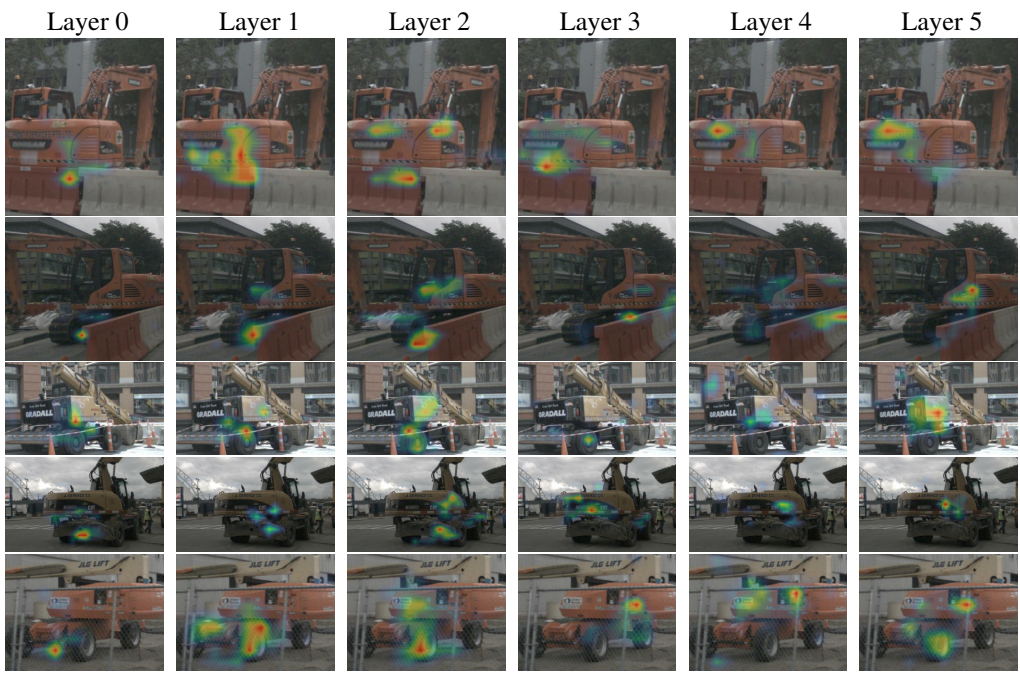}
            \end{subfigure}
        }
        
    }
\label{fig:ex_construction}
\caption{Raw cross-attention $\mathbb{E}_h( A_{CR} )^+$ examples for class \texttt{construction vehicles}.}
\end{figure}

\clearpage
\section{Examples with Camera FOV Overlap}\label{appendix:examples_fov_overlap}
\begin{figure}[hbt!]
    \begin{minipage}{0.5\textwidth}
        \centering
        \scriptsize FRONT CAMERA 56\%
    \end{minipage}
    \begin{minipage}{0.5\textwidth}
        \centering
        \scriptsize FRONT RIGHT CAMERA 42\%
    \end{minipage}
    \includegraphics[width=1\textwidth]{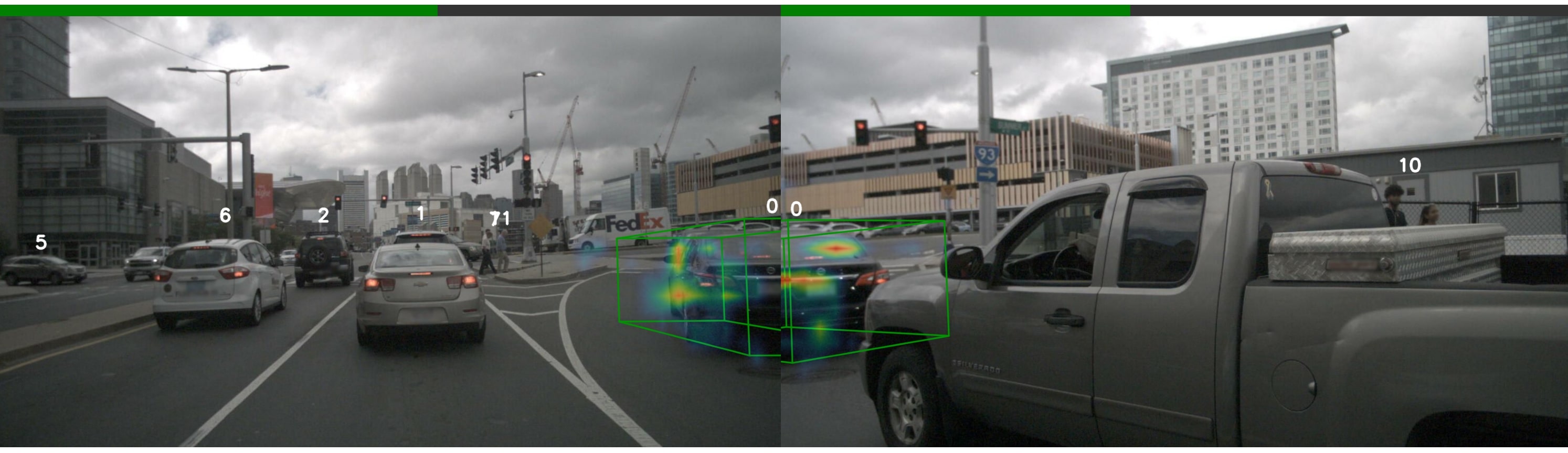}
    \begin{minipage}{0.5\textwidth}
        \centering
        \scriptsize FRONT CAMERA 71\%
    \end{minipage}
    \begin{minipage}{0.5\textwidth}
        \centering
        \scriptsize FRONT RIGHT CAMERA 23\%
    \end{minipage}
    \includegraphics[width=1\textwidth]{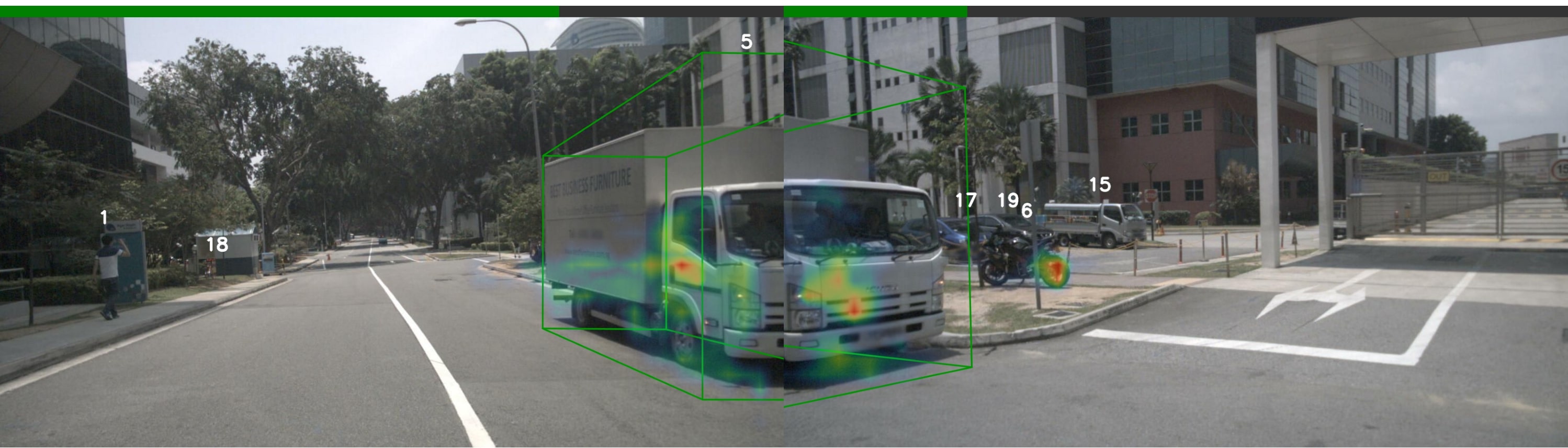}
    \begin{minipage}{0.5\textwidth}
        \centering
        \scriptsize FRONT CAMERA 24\%
    \end{minipage}
    \begin{minipage}{0.5\textwidth}
        \centering
        \scriptsize FRONT RIGHT CAMERA 75\%
    \end{minipage}
    \includegraphics[width=1\textwidth]{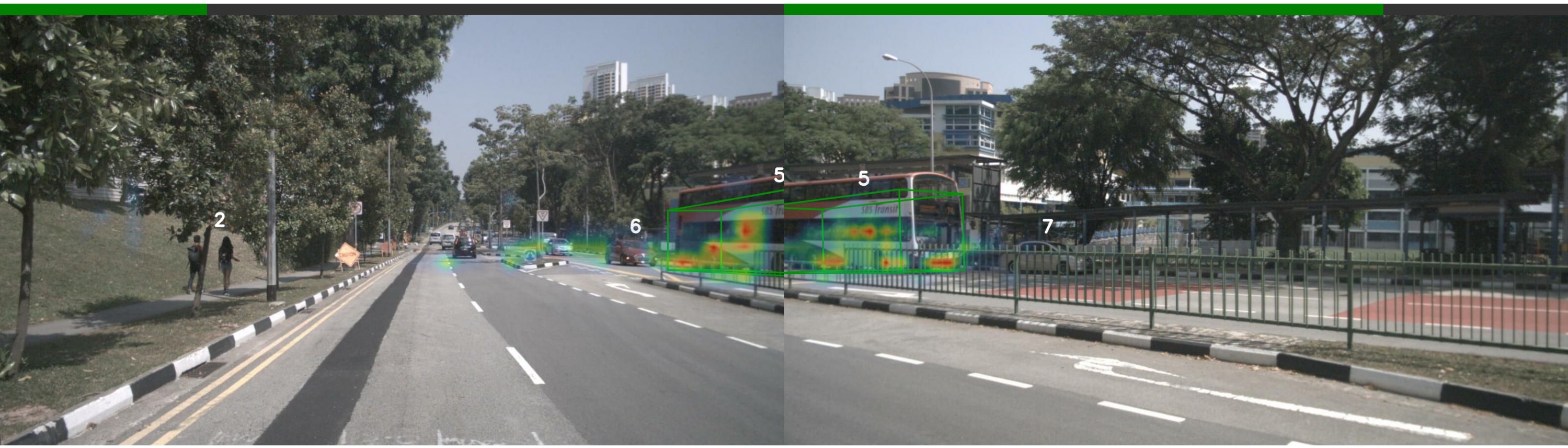}
    \begin{minipage}{0.5\textwidth}
        \centering
        \scriptsize FRONT LEFT CAMERA 80\%
    \end{minipage}
    \begin{minipage}{0.5\textwidth}
        \centering
        \scriptsize FRONT CAMERA 17\%
    \end{minipage}
    \includegraphics[width=1\textwidth]{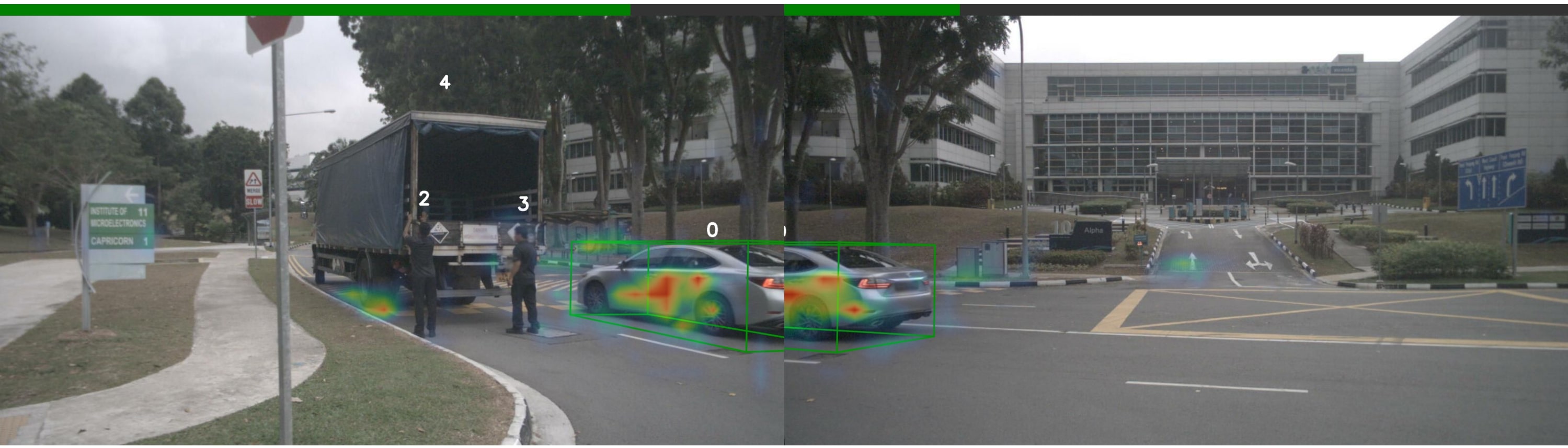}
    \caption{Raw cross-attention examples for objects (green OBB) that lie in the overlapping FOV of two cameras. In each example, a single query is used to generate the saliency maps for all camera images. Attention can be observed on the object on both overlapping camera images. We further approximated the contribution of each camera for a detection by computing the faction of attention for each camera, indicated by the green bar and the percentage data.}
    \label{fig:ex_with_camera_fov_overlap}
\end{figure}

The nuScenes dataset \cite{Caesar2020} was recorded with a multi-camera setup consisting of six cameras which have a partial overlap in their FOVs. The SpatialDETR transformer architecture projects each query onto all camera frames, enabling a \emph{global attention} on all input images. This end-to-end approach generates a single detection for objects that appear in two or more cameras at the same time. In such cases, the attention is distributed on the overlapping parts of the object, as shown in Fig. \ref{fig:ex_with_camera_fov_overlap}.

       

        

\end{document}